# ABSTRACT

**Title of Dissertation:** Development of ICA and IVA Algorithms with Application to Medical Image Analysis

Zois Boukouvalas, Doctor of Philosophy, 2017

**Dissertation directed by:**  Dr. Tülay Adalı, Distinguished Professor
Department of Computer Science and
Electrical Engineering

Blind source separation (BSS) is an active area of research due to its applicability to a variety of problems, especially when there is a little known about the observed data. Applications where BSS has been successfully utilized include the analysis of medical imaging data, such as functional magnetic resonance imaging (fMRI) data, detection of specific targets in video sequences or multi-spectral remote sensing data, among many others. Independent component analysis (ICA) is a widely used BSS method that can uniquely achieve source recovery, subject to only scaling and permutation ambiguities, through the assumption of statistical independence on the part of the latent sources. Independent vector analysis (IVA) extends the applicability of ICA by jointly decomposing multiple datasets through the exploitation of the dependencies across datasets. Though both ICA and IVA algorithms cast in the maximum likelihood (ML) framework enable the use of all available statistical information—forms of diversity—in reality, they often deviate from their theoretical optimality properties due to improper estimation of the probability density function (PDF). This motivates the development of flexible ICA and IVA algorithms that closely adhere to the underlying statistical description of the data. Although it is attractive to let the data "speak" and hence minimize the assumptions, important prior information about the data, such as sparsity, is usually available. If incorporated into the ICA model, use of this additional information can relax the independence assumption, resulting in an improvement in the overall separation performance. Therefore, the development of a unified mathematical framework that can take into account both statistical independence and sparsity is of great interest.

In this dissertation, we first introduce a flexible ICA algorithm that uses an effective PDF estimator to accurately capture the underlying statistical properties of the data, yielding superior separation performance and maintaining the desirable optimality of ML estimation. We then discuss several techniques to accurately estimate the parameters of the multivariate generalized Gaussian distribution, and how to integrate them into the IVA model and derive a class of flexible IVA algorithms that take both second-order statistics and higher-order statistics into account. Finally, we provide a mathematical framework that enables direct control over the influence of statistical independence and sparsity, and use this framework to develop an effective ICA algorithm that can jointly exploit these two forms of diversity. Hence, by increasing the flexibility of ICA and IVA algorithms and by enriching the models through the incorporation of reliable priors, we enhance the capabilities of ICA and IVA and, thus, enable their application to many problems. We demonstrate the effectiveness of the proposed ICA and IVA algorithms using numerical examples as well as fMRI-like data.

# Development of ICA and IVA Algorithms

# with Application to Medical Image Analysis

by

Zois Boukouvalas

Dissertation submitted to the Faculty of the Graduate School
of the University of Maryland in partial fulfillment
of the requirements for the degree of
Doctor of Philosophy
2017

# TABLE OF CONTENTS

# LIST OF FIGURES

# LIST OF TABLES

# LIST OF ABBREVIATIONS

| | |
|---|---|
| BFGS | Broyden, Fletcher, Goldfarb, and Shanno optimization algorithm |
| BSS | blind source separation |
| C-ICA | constrained independent component analysis |
| CNR | contrast to noise ratio |
| DL | dictionary learning |
| EEG | electroencephalography |
| EFICA | the efficient variant of FastICA algorithm |
| EMK | entropy maximization with kernels |
| FastICA | fast independent component analysis |
| FastICAskew | fast independent component analysis with skewed nonlinearity |
| FastICAtanh | fast independent component analysis with tanh nonlinearity |
| fMRI | functional magnetic resonance imaging |
| FP | fixed point |
| FS | Fisher scoring |
| GGD | generalized Gaussian distribution |
| HOS | higher-order statistics |
| i.i.d. | independent and identically distributed |
| ICA | independent component analysis |
| ICA-EBM | ICA by entropy bound minimization |
| ICA-EMK | ICA by entropy maximization with kernels |
| Infomax | information maximization |
| Infomax-BFGS | information maximization using BFGS optimization technique |
| Infomax-NG | information maximization using natural gradient optimization technique |
| ISI | intersymbol-interference |
| ISR | interference to source ratio |
| IVA | independent vector analysis |
| IVA-A-GGD-MLFS | IVA with adaptive MGGD using maximum likelihood FS |
| IVA-A-GGD-MoM | IVA with adaptive MGGD using method of moments |
| IVA-A-GGD-RAFP | IVA with adaptive MGGD using Riemannian averaged FP |
| IVA-G | IVA with multivariate Gaussian |
| IVA-GGD | IVA with multivariate generalized Gaussian distribution |
| IVA-L | IVA with multivariate Laplacian |
| JDIAG-HOS | joint diagonalization fourth-order cumulant matrices |
| JDIAG-SOS | joint diagonalization via second-order statistics |
| KDE | kernel density estimation |
| KL | Kullback-Leibler |
| kNN | k-nearest neighbors |
| MDL | minimum description length |
| MGGD | multivariate generalized Gaussian distribution |
| MI | mutual information |
| ML | maximum likelihood |
| ML-FP-eps | maximum likelihood fixed point using $\epsilon$ perturbation |
| MN | maximization of neg-entropy |
| MoM | method of moments |
| NPAIRS | nonparametric, prediction, activation, influence, reproducibility, resampling |
| PCA | principal component analysis |
| PDF | probability density function |
| RADICAL | robust, accurate, direct independent components analysis algorithm |
| RA-FP | Riemannian averaged fixed point |
| SCA | sparse component analysis |
| SCV | source component vector |
| SNP | single nucleotide polymorphism |
| SOS | second-order statistics |
| SparseICA-EBM | sparse independent component analysis by entropy bound minimization |

Chapter 1

# INTRODUCTION

## 1.1 Motivation

In many real-world applications, it is beneficial to summarize the observed data through the use of a latent factor model [30, 43]. Since, in the majority of applications, little is known about the processes underlying the generation of the factors, it is important to minimize the assumptions placed on the data in order to avoid potentially biasing the solution. This motivates the development of the field of blind source separation (BSS), which extracts summary factors with few assumptions placed on the data, generally through the use of a generative model. Because of this ability, there are numerous applications where BSS techniques are used, including: analysis of medical imaging data [68, 69], detection of targets in video sequences [16] or sets of images such as multi-spectral remote sensing data, separation of audio [83], feature extraction from images [45], and in (antenna) array processing [25], among many others.

One of the most widely used BSS techniques is independent component analysis (ICA) [1]. The reason for this popularity is the fact that through only the assumption of statistical independence on the part of the latent sources, ICA is able to uniquely identify the true latent sources subject to only scaling and permutation ambiguities. A natural way to achieve this is through the use of maximum likelihood (ML) estimation, which enables one to take into account all statistical properties of the data—forms of diversity. Additionally, formulating ICA under an ML framework provides many theoretical advantages enabling the study of asymptotic optimality of the estimator, derivation of a lower bound on variance (Cramèr-Rao lower bound), and identifiability conditions [1, 30]. Most ICA algorithms can be derived as special cases of the ML cost function [1, 2]. However, in many applications, knowledge of the underlying probability density function (PDF) of the latent sources is generally unknown. Algorithms that utilize a fixed model for the underlying distribution of the latent sources or a simple model, *i.e.*, one that is not flexible, can yield poor separation performance when the data deviates from the assumed model [20]. In such cases, these algorithms also diverge from the desirable optimality conditions of the ML estimation. This motivates the development of flexible ICA algorithms that use a PDF estimator to closely adhere to the underlying statistical description of the data, thus yielding superior separation performance while maintaining the desirable optimality of ML estimation.

Although ICA is one of the most commonly used BSS techniques, it can only decompose a single dataset. However, in many applications multiple sets of data are gathered about the same phenomenon, thus motivating the development of methods that jointly factorize this multi-set data. This has driven the development of independent vector analysis (IVA), a recent generalization of ICA to multiple datasets that can achieve improved performance over performing ICA on each dataset separately by exploiting dependencies across datasets. Applications where IVA is a natural solution include: solution to the convolutive ICA problem [55], video surveillance [15], and analysis of brain activity using medical image data collected from multiple subjects [1, 66]. Like ICA, IVA can be formulated in a ML framework such that all available types of diversity are taken into account simultaneously, as is the case for ICA, through the use of density general models for the latent multivariate sources [1, 5]. The multivariate generalized Gaussian distribution (MGGD) provides an effective model for the latent multivariate sources in IVA and accounts for many forms of diversity such as second-order statistics (SOS) and higher order statistics (HOS). However, its performance in terms of its separation power highly depends on the estimation of the source parameters. Therefore, techniques to estimate the MGGD parameters, the symmetric positive definite scatter matrix and the shape parameter, and successfully integrating them into IVA is of considerable interest.

While it is attractive to let the data "speak" and hence minimize the assumptions, in practice, important prior information about the underlying latent sources is usually available. A widely used approach for incorporating prior information into the ICA framework is through the use of constrained independent component analysis (C-ICA) [64], which incorporates prior information using equality and inequality constraints under a Lagrangian framework. Such prior information can be about the task in functional magnetic resonance imaging fMRI analysis and can be included as constraints on the mixing matrix columns [21, 80, 96] or spatial maps [63, 64, 85]. While this approach is practical, such constraints have to be in an exact functional form, something that is not always available in practice. Another form of prior information that can be considered are natural properties of the data, such as sparsity. If incorporated into the ICA model, they can relax the independence assumption, resulting in an improvement in the overall separation performance. This motivates the development of a mathematical framework that enables direct control over the influence that independence and sparsity have on the result and using this framework to generate a powerful algorithm that takes both sparsity and independence into account.

To summarize, the key issue that enables the application of ICA and IVA algorithms to many problems is the development of effective models for the underlying source densities, their estimation and efficient utilization of prior information. The objective of this dissertation is the development of effective ICA and IVA algorithms that can face these challenges and study of their application to fMRI data.

## 1.2 Contributions

In this dissertation, we address the challenges stated above and specifically make the following contributions:

**Flexible Probability Estimation Technique for ICA**

In order to achieve the optimality conditions for ICA using the ML framework, exact knowledge of the true PDF of the latent sources is required. However, this information is usually not available in most real-world applications. Algorithms that utilize a fixed or simple model for the underlying distribution of the latent sources can yield poor separation performance when the data significantly deviates from the assumed model. We

- Develop a new and efficient ICA algorithm based on entropy maximization with kernels, (ICA-EMK), which uses both global and local measuring functions as constraints to dynamically estimate the PDF of the sources with reasonable complexity;

- Derive an optimization framework, enabling parallel implementations on multi-core computers;

- Demonstrate the superior performance of ICA-EMK over competing ICA algorithms using simulated as well as real-world data.

**MGGD Parameter Estimation and Efficient Integration to IVA**

The MGGD has been an attractive solution to many signal processing problems due to its simple yet flexible parametric form, which requires the estimation of only a few parameters, *i.e.*, the scatter matrix and the shape parameter. MGGD provides an effective model for IVA as well as for modeling the latent multivariate variables–sources–and the performance of the IVA algorithm highly depends on the estimation of the source parameters. We

- Develop a new and efficient ML estimation technique based on the Fisher scoring (FS) that estimates both the shape parameter and the scatter matrix;

- Develop a new fixed point (FP) algorithm, called Riemannian averaged FP (RA-FP) that accurately estimates the scatter matrix for any positive value of the shape parameter;

- Present a theoretical justification of the convergence of RA-FP;

- Derive a new IVA algorithm, IVA with adaptive MGGD, that estimates the shape parameter and scatter matrix jointly and takes into account both second and higher-order statistics.

**Incorporation of Prior Information into the ICA Model**

Though ICA has proven powerful in many applications, complete statistical independence can be too restrictive an assumption in many applications. Additionally, important prior information about the data, such as sparsity, is usually available. Sparsity is a natural property of the data, a form of diversity, which, if incorporated into the ICA model, can relax the independence assumption, resulting in an improvement in the overall separation performance. We

- Provide a mathematical framework for blind source separation that enables direct control over the relative influence of independence and sparsity;

- Develop a new ICA algorithm, sparse ICA by entropy bound minimization (SparseICA-EBM) through the direct exploitation of sparsity;

- Demonstrate that the proposed algorithm inherits all the advantages of ICA-EBM, namely its flexibility, though with enhanced performance due to the exploitation of sparsity and allow the user to balance the roles of independence and sparsity;

- Study the synergy of independence and sparsity through simulations on synthetic as well as fMRI-like data;

- Explore the trade-offs between independence and sparsity in the ICA optimization framework and provide a guidance on how to balance these two objectives in real world applications where the ground truth is not available.

## 1.3 Overview of dissertation

This dissertation is organized as follows.

In Chapter 2, we provide the necessary mathematical background for the ICA and IVA model and discuss popular ICA and IVA algorithms. We then motivate and derive a decoupling procedure which allows optimization of vectors rather than matrices, providing benefits such as integration of flexible PDF estimation techniques in the ICA and IVA models, incorporation of constraints in the ICA/IVA framework, as well as the development and implementation of parallel ICA/IVA algorithms.

In Chapter 3, we discuss several PDF estimation techniques and present a flexible ICA algorithm that is based on the maximum entropy principle along with its pseudo-code. We discuss its parallel implementation and demonstrate its effectiveness on both simulated as well as well as mixtures of face images.

In Chapter 4, introduce two ML estimation techniques, to estimate shape parameter and the scatter matrix of an MGGD. Using these two techniques, we derive a class of IVA algorithms that estimate shape parameter and scatter matrix jointly and take into account both SOS and HOS. We show that the new IVA algorithms provide desirable performance by comparing their performance with those of competing algorithms.

Chapter 5 presents a mathematical framework that enables direct control over the influence of independence and sparsity and then apply the proposed framework to the development of an effective ICA algorithm that can jointly exploit this two forms of diversity. Using several evaluation metrics we demonstrate the superior performance of the new al-gorithm on simulated sparse sources as well as simulated fMRI sources .

Chapter 6 provides a summary of the results of the dissertation and suggestions for future work.

# Chapter 2

# ICA AND IVA: THEORY AND ALGORITHMS

ICA is a powerful method for BSS that can achieve nearly perfect source recovery through the assumption of statistical independence of the latent sources and by making use of different statistical properties—forms of diversity—of the data. In many applications, a joint analysis of multiple datasets needs to be performed, such as medical imaging data from multiple subjects or at different conditions, motivating the development of methods that can exploit the complementary information across these multivariate datasets. One such method is IVA, a recent generalization of ICA to multiple datasets that has been shown to achieve better performance than performing ICA separately on each dataset by taking the dependence across datasets into account. We begin this chapter by formulating the BSS problem and discuss the role of diversity in BSS. Then, we present ICA and IVA and derive the ML objective functions for both models and show their relation to the mutual information (MI) objective function as the number of samples approaches infinity. The connection to the MI objective function provides a framework that enables the exploitation of multiple forms of diversity, while at the same time enjoying all the theoretical advantages of ML theory. Finally, we discuss popular ICA and IVA algorithms that fall under the maximum likelihood umbrella and stress the key issues that enable the application of ICA and IVA algorithms to many problems.

## 2.1 Blind Source Separation

BSS is an active area of research in statistical signal processing due to its numerous applications. As we discuss in Chapter 1, some examples include analysis of medical imaging data, wireless communications, and image processing. The objective of BSS methods is to decompose a matrix containing a set of observations into the product of a mixing matrix and a matrix of latent sources.

To mathematically formulate the BSS problem, let $\mathbf{X} \in \mathbb{R}^{M \times V}$ denote the observation matrix where $M$ denotes the number of observation vectors and $V$ denotes the number of samples. The noiseless BSS generative model is given by

$$\mathbf{X} = \mathbf{AS}, \tag{2.1}$$

where $\mathbf{A} \in \mathbb{R}^{M \times N}$ is the mixing matrix and $\mathbf{S} \in \mathbb{R}^{N \times V}$ is the matrix that contains the source signals. However, without the exploitation of any prior knowledge about the data, most typically its statistical properties—forms of diversity—the matrix factorization problem is ill-posed. This can been seen, for example, since for any invertible matrix $\mathbf{T} \in \mathbb{R}^{N \times N}$, it always holds that

$$\mathbf{X} = \mathbf{AS} = (\mathbf{AT})(\mathbf{T}^{-1}\mathbf{S}).$$

Some forms of diversity that could be imposed in the mixing and/or the source matrix are: statistical independence of the sources, sparsity, correlation of sources across datasets, sample-to-sample dependence [52], or sparsity [28, 37, 67, 101]. Independence, in particular [9, 12, 44, 55, 79], is a reasonable assumption in many real world applications and its assumption enables source estimation subject to only scaling and permutation ambiguity.

## 2.2 Independent Component Analysis

One of the most widely used methods for solving the BSS problem (2.1) is ICA and its basic assumption is that the source signals are statistically independent. By rewriting (2.1) using random vector notation, we have

$$\mathbf{x}(v) = \mathbf{A}\mathbf{s}(v), \quad v = 1, \ldots, V,$$

where $v$ is the sample index, $\mathbf{s}(v) \in \mathbb{R}^N$ are the unknown source signals, $\mathbf{x}(v) = [x_1(v), \ldots, x_M(v)]^\top \in \mathbb{R}^M$ are the mixtures and $\mathbf{A} \in \mathbb{R}^{N \times N}$ is the mixing matrix. A common case in many applications such as the analysis of fMRI data, is the overdetermined one ($M > N$), which can be reduced to the case where $M = N$ using dimensionality reduction, generally through the performance of principal component analysis (PCA). Under the assumption that the sources $s_n(v)$, $1 \leq n \leq N$ in $\mathbf{s}(v) = [s_1(v), \ldots, s_N(v)]^\top$ are statistically independent and making use of different properties of the signals, the goal in ICA is to estimate a demixing matrix $\mathbf{W} \in \mathbb{R}^{N \times N}$ to yield maximally independent source estimates $\mathbf{y}(v) = \mathbf{W}\mathbf{x}(v)$.

### 2.2.1 ICA Objective Functions

ML theory possesses many theoretical advantages allowing the study of asymptotic normality, consistency, and efficiency of an estimator. Thus, casting ICA under the ML umbrella enables the derivation of the CRLB and the determination of identifiability conditions [1, 30] for the ICA model. In this section, we derive the ICA ML objective function and connect it to the objective function that is related to the mutual information (MI) of the estimated sources. We show that both objective functions yield equivalent solutions at their optimum.

For the purpose of this chapter and for the rest of our thesis we assume that the samples of each source are independent and identical distributed (i.i.d). The parameter that needs to be estimated in ICA is the inverse of the mixing matrix $\mathbf{A}$, *i.e.*, $\mathbf{W} = \mathbf{A}^{-1}$ subject to only scaling and permutation ambiguities. The ML objective function is given by

$$\begin{aligned}
\mathcal{L}_{ICA}(\mathbf{W}) &= \log\left(\prod_{v=1}^{V} p_{\mathbf{x}}(\mathbf{x}(v))\right) \\
&= \sum_{v=1}^{V} \log(p_{\mathbf{s}}(\mathbf{W}\mathbf{x}(v))|\det(\mathbf{W})|) \\
&= \sum_{v=1}^{V} \log(p_{\mathbf{s}}(\mathbf{W}\mathbf{x}(v))) + V\log(|\det(\mathbf{W})|) \\
&= \sum_{v=1}^{V} \left[\sum_{n=1}^{N} \log p_{s_n}(\mathbf{w}_n^\top \mathbf{x}(v))\right] + V\log|\det(\mathbf{W})| \\
&= V\left(\frac{1}{V}\sum_{v=1}^{V}\left[\sum_{n=1}^{N} \log p_{s_n}(\mathbf{w}_n^\top \mathbf{x}(v))\right] + \log|\det(\mathbf{W})|\right),
\end{aligned}$$

where $p(\mathbf{w}_n^\top \mathbf{x})$ is the PDF of the estimated random variable $y_n = \mathbf{w}_n^\top \mathbf{x}$. Since $V \neq 0$, and maximization is scale invariant the ML objective function is simplified to

$$\mathcal{L}_{ICA}(\mathbf{W}) = \frac{1}{V}\sum_{v=1}^{V}\left[\sum_{n=1}^{N} \log p_{s_n}(\mathbf{w}_n^\top \mathbf{x}(v))\right] + \log|\det(\mathbf{W})|. \qquad (2.2)$$

Using the mean ergodic theorem and as $V \to \infty$, (2.2) becomes

$$\begin{aligned}
\mathcal{L}_{ICA}(\mathbf{W}) &= E\left\{\sum_{n=1}^{N} \log p(y_n)\right\} + \log|\det(\mathbf{W})| \\
&= \sum_{n=1}^{N} E\{\log p(y_n)\} + \log|\det(\mathbf{W})|. \qquad (2.3)
\end{aligned}$$

due to the linearity of the expected value. From an optimization perspective, the term $\log(|\det(\mathbf{W})|)$ acts as a regularization parameter and prevents the minimization of the cost function by simply scaling its first term by any non-zero constant. Since there are no constraints imposed to the optimization problem, apart from $\mathbf{W}$ being nonsingular, the solution space is defined by the space of all invertible matrices.

Another approach to define an ICA objective function is by the MI among the estimated sources and is defined by the Kullback Leibler (KL)-distance between the joint source density and the product of the marginal estimated source densities. Thus, by using the fact that $p_{\mathbf{s}}(\mathbf{W}\mathbf{x}) = p_{\mathbf{x}}(\mathbf{x})|\det(\mathbf{W})|^{-1}$, the MI objective function is given by

$$\begin{aligned} J_{ICA}(\mathbf{W}) &= E\left\{-\log\left[\frac{p_{s_1}(y_1)p_{s_2}(y_2)\cdots p_{s_N}(y_N)}{p_{s_1 s_2 \ldots s_N}(y_1, y_2, \ldots, y_N)}\right]\right\} \\ &= E\left\{-\sum_{n=1}^{N}\log p_{s_n}(y_n)\right\} + E\left\{\log p_{\mathbf{s}}(\mathbf{y})\right\} \\ &= \sum_{n=1}^{N} H(y_n) - H(\mathbf{y}) \\ &= \sum_{n=1}^{N} H(y_n) - \log|\det(\mathbf{W})| - H(\mathbf{x}), \end{aligned} \qquad (2.4)$$

where the terms $H(y_n)$ and $H(\mathbf{x})$ are the (differential) entropy of the source estimates and the mixtures, respectively. Note that the term $H(\mathbf{x})$ is independent of $\mathbf{W}$ and can be treated as a constant. It is clear that minimizing (2.4) is not a straightforward since there is no access to the true underlying probability density function (PDF) of each estimated source. Therefore, if $\hat{p}_{s_n}(y_n)$ denotes the PDF of the $n$th source estimate, then

$$\begin{aligned} H(y_n) &= -\int_{-\infty}^{\infty} p_{s_n}(y_n) \log p_{s_n}(y_n)\, dy_n \\ &= -\int_{-\infty}^{\infty} p_{s_n}(y_n) \log \frac{p_{s_n}(y_n)}{\hat{p}_{s_n}(y_n)}\, dy_n - \int_{-\infty}^{\infty} p_{s_n}(y_n) \log \hat{p}_{s_n}(y_n)\, dy_n \\ &= -f(p_{s_n}(y_n), \hat{p}_{s_n}(y_n)) - E\{\log \hat{p}_{s_n}(y_n)\}, \end{aligned} \qquad (2.5)$$

where $f(p_{s_n}(y_n), \hat{p}_{s_n}(y_n))$ denotes the KL distance between the $n$th estimated and the true source PDF. The expression in (2.5) indicates that maximization of (2.3) is equivalent to the minimization of the MI objective function, as long as the assumed model PDF matches the true latent source PDF *i.e.,* $f(p_{s_n}(y_n), \hat{p}_{s_n}(y_n)) = 0$. Note that for the rest of this section as well as for the definition of the IVA objective functions, we assume that the source PDF is known.

### 2.2.2 ICA Algorithms

Using the MI objective function (2.4) the derivative with respect to (w.r.t) $\mathbf{W}$ is derived,

$$\frac{\partial J_{ICA}(\mathbf{W})}{\partial \mathbf{W}} = \sum_{n=1}^{N} \frac{\partial H(y_n)}{\partial \mathbf{W}} - \frac{\partial \log|\det(\mathbf{W})|}{\partial \mathbf{W}}$$

$$= -\sum_{n=1}^{N} E\left\{\frac{\partial \log p_{s_n}(y_n)}{\partial \mathbf{W}}\right\} - \mathbf{W}^{-\top}$$

$$= -\sum_{n=1}^{N} E\left\{\frac{\partial \log p_{s_n}(y_n)}{\partial y_n} \frac{\partial y_n}{\partial \mathbf{W}}\right\} - \mathbf{W}^{-\top}. \tag{2.6}$$

Using the fact that the derivative of the second term in the expected value is given by

$$\left[\frac{\partial y_n}{\partial \mathbf{W}}\right]_{j,i} = \frac{\partial (\mathbf{w}_n)^\top \mathbf{x}}{\partial w_{j,i}} = x_i \delta_{j,n}, \tag{2.7}$$

where $d_{j,n}$ is the Kronecker delta function. The derivative of the MI objective function is given by

$$\frac{\partial J_{ICA}(\mathbf{W})}{\partial \mathbf{W}} = E\{\boldsymbol{\phi}\mathbf{x}^\top\} - \mathbf{W}^{-\top}, \tag{2.8}$$

where $\boldsymbol{\phi} = -\left[\frac{\partial \log p_{s_1}(y_1)}{\partial y_1}, \ldots, \frac{\partial \log p_{s_N}(y_N)}{\partial y_N}\right]^\top$ and is called the score function. Thus the matrix $\mathbf{W}$ can be updated using the steepest descent,

$$(\mathbf{W})^{\text{new}} \leftarrow (\mathbf{W})^{\text{old}} - \gamma \frac{\partial J_{ICA}(\mathbf{W})}{\partial \mathbf{W}}, \tag{2.9}$$

where $\gamma$ is a scalar step size and can be fixed to a small number or a line search algorithm can be used to determine the largest $\gamma$ that results in a decrease in the ICA cost function. The update rule in (2.9) requires the inversion of $\mathbf{W}$ at each ICA iteration. This poses several computational issues: first, calculating the inverse of a matrix significantly increases the computational complexity of an algorithm, especially as the number of sources increases, second, if **W** is ill-conditioned its inversion will introduce computational errors.

These issues can be avoided by using relative/natural gradient updates as proposed in [24, 27]. The natural gradient is the gradient of (2.6) post-multiplied by $\mathbf{W}^\top\mathbf{W}$. Therefore, (2.9) becomes

$$(\mathbf{W})^{\text{new}} \leftarrow (\mathbf{W})^{\text{old}} - \gamma(E\{\boldsymbol{\phi}\mathbf{x}^\top\}(\mathbf{W}^\top)^{\text{old}} - \mathbf{I})(\mathbf{W})^{\text{old}}. \quad (2.10)$$

Since the $\mathbf{W}^\top\mathbf{W}$ is positive definite, the relative/natural gradient does not alter the stability of the solution [4].

Solutions for the estimation of the components of the score function include parametric, non-parametric, and semi-parametric approaches. FastICA [42], efficient fast ICA (EFICA) [50], and information maximization (Infomax) [11], are based on a parametric approach and use a fixed nonlinearity or model for the underlying distribution of the sources, making them computationally attractive. However, their separation performance suffers when the density of the true sources deviates significantly from the assumed model. Robust, accurate, direct ICA (RADICAL) [54] is a nonparametric ICA algorithm, which uses estimates of entropy. However, nonparametric methods are practically difficult due to the parameter selection that is required and are computationally demanding when number of samples increases. ICA by entropy bound minimization (ICA-EBM) [57] is based on a semi-parametric approach and provides flexible density matching through use of four measuring functions based on the maximum entropy principle. Four measuring functions are used for calculating the entropy bound, but the associated maximum entropy density is limited to bimodal, symmetric or skewed, heavy-tailed or not heavy-tailed distributions, which might be limited to scenarios where the PDF of the latent sources is unknown and complicated.

## 2.3 Independent Vector Analysis

In many practical applications, multiple data sets with dependence among them need to be analyzed. Such applications include the analysis of fMRI and EEG data from multiple subjects or different conditions. IVA generalizes the ICA problem by allowing for full exploitation of the dependence across multiple datasets as well as the properties that ICA can make use of, which leads to improved performance beyond what is achievable by applying separately to each dataset. Additionally, IVA automatically aligns dependent sources across the datasets, thus bypassing the need of a second permutation correction algorithm for the task.

The IVA problem is mathematically formulated similar to the ICA one, except that now we have $K$ datasets $\mathbf{x}^{[k]}$, $k = 1, ..., K$ where each dataset is a linear mixture of $N$ statistically independent sources. Using random vector notation under the assumption that samples are i.i.d the noiseless IVA model is given by

$$\mathbf{x}^{[k]} = \mathbf{A}^{[k]} \mathbf{s}^{[k]}, \quad k = 1, ..., K,$$

where $\mathbf{A}^{[k]} \in \mathbb{R}^{N \times N}$, $k = 1, ..., K$ are invertible mixing matrices and $\mathbf{s}^{[k]} = [s_1^{[k]}, ..., s_N^{[k]}]^\top$ is the vector of latent sources for the $k$th dataset. In the IVA model, the components within each $\mathbf{s}^{[k]}$ are assumed to be independent, while at the same time, the IVA model allows for dependence across corresponding components of $\mathbf{s}^{[k]}$ in multiple datasets. This additional type of diversity that IVA can take into account, comes from the definition of the source component vector (SCV) that is defined by vertically concatenating the $n$th source from each of the $K$ datasets and is denoted as

$$\mathbf{s}_n = [s_n^{[1]}, ..., s_n^{[K]}]^\top, \quad (2.11)$$

where $\mathbf{s}_n$ is a $K$-dimensional random vector. In contrast to ICA, the goal in IVA is to estimate $K$ demixing matrices to yield source estimates $\mathbf{y}^{[k]} = \mathbf{W}^{[k]}\mathbf{x}^{[k]}$, such that each SCV is maximally independent of all other SCVs. An example that demonstrates the additional diversity that the IVA makes use of is fMRI analysis. Usually in fMRI analysis there are multiple subjects and one would expect the $n$th activation maps of the different subjects that form the $n$th SCV to be statistical dependent. A graphical illustration of SCV using matrix notation is shown in Figure 2.1. It is worth mentioning, that IVA does not require dependence across datasets to exist. In the case where this form of diversity does not exist, IVA reduces to individual ICAs on each dataset.

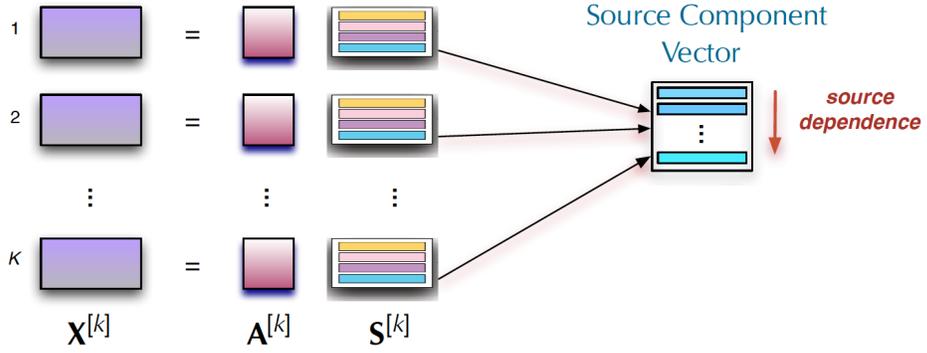

FIG. 2.1. Illustration of an SCV and source vectors.

### 2.3.1 IVA Objective Function

Similar to ICA, IVA can be formulated in a ML framework. However, the optimization parameter is not just a single demixing matrix $\mathbf{W}$ as in the ICA case, but a set of demixing matrices $\mathbf{W}^{[1]}, \ldots, \mathbf{W}^{[K]}$, which can be collected into a three dimensional array $\mathcal{W} \in \mathbb{R}^{N \times N \times K}$. The ML objective function for IVA is given by

$$\mathcal{L}_{IVA}(\mathcal{W}) = \sum_{n=1}^{N} E\left\{\log p(\mathbf{y}_n)\right\} + \sum_{k=1}^{K} \log |\det(\mathbf{W})^{[k]}|, \tag{2.12}$$

where $\mathbf{y}_n$ is the *n*th estimated random vector and $p(\mathbf{y}_n)$ denotes its multidimensional PDF. As in the ICA case, maximization of (2.12) is asymptotically equivalent to the minimization of the MI objective function, as long as the assumed model PDF matches the true latent source PDF. The MI objective function for the IVA model is given by

$$J_{IVA}(\mathcal{W}) = \sum_{n=1}^{N} H(\mathbf{y}_n) - \sum_{k=1}^{K} \log \left| \det \left( \mathbf{W}^{[k]} \right) \right| - H(\mathbf{x}^{[1]}, ..., \mathbf{x}^{[K]}), \qquad (2.13)$$

where $H(\mathbf{y}_n)$ denotes the entropy of the *n*th SCV and the term $H(\mathbf{x}^{[1]}, ..., \mathbf{x}^{[K]})$ can be treated as a constant. To see the role of the additional form of diversity that we take into account using the IVA framework, (2.13) can be written as

$$J_{IVA}(\mathcal{W}) = \sum_{n=1}^{N} \left( \sum_{k=1}^{K} H(y_n^k) - I(\mathbf{y}_n) \right) - \sum_{k=1}^{K} \log \left| \det \left( \mathbf{W}^{[k]} \right) \right| - H(\mathbf{x}^{[1]}, ..., \mathbf{x}^{[K]}), \qquad (2.14)$$

where the term $I(\mathbf{y}_n)$ denotes the mutual information within the *n*th SCV and minimization of (2.14) automatically increases the mutual information within the components of an SCV, revealing how IVA exploits the additional form of diversity into account. It can be seen that without the mutual information term, the objective function (2.14) is equivalent to performing independent ICA separately on each dataset.

### 2.3.2 IVA Algorithms

Using the IVA MI objective function the derivative w.r.t. to each of the demixing matrices is derived similar to the one for ICA and is given by

$$\frac{\partial J_{IVA}(\mathcal{W})}{\partial \mathbf{W}^{[k]}} = E \left\{ \boldsymbol{\phi}^{[k]} (\mathbf{x}^{[k]})^\top \right\} - (\mathbf{W}^{[k]})^{-\top}, \qquad (2.15)$$

where $\boldsymbol{\phi}^{[k]} = -\left[ \frac{\partial \log p_{s_1}(y_1)}{\partial y_1^{[k]}}, \ldots, \frac{\partial \log p_{s_N}(y_N)}{\partial y_N^{[k]}} \right]^\top$. Thus, each of the *K* demixing matrices is updated using

$$(\mathbf{W}^{[k]})^{\text{new}} \leftarrow (\mathbf{W}^{[k]})^{\text{old}} - \gamma \frac{\partial J_{IVA}(\mathcal{W})}{\partial \mathbf{W}^{[k]}}, \qquad (2.16)$$

where $\gamma$ is the step size. Similar to the ICA case, the *k*th estimate of the demixing matrix needs to be inverted at each IVA iteration. Thus by using relative/natural gradient updates as in [6], (2.16) becomes

$$(\mathbf{W}^{[\mathbf{k}]})^{\text{new}} \leftarrow (\mathbf{W}^{[\mathbf{k}]})^{\text{old}} - \gamma(E\{\boldsymbol{\phi}\mathbf{x}^\top\}(\mathbf{W}^{[\mathbf{k}]\top})^{\text{old}} - \mathbf{I})(\mathbf{W}^{[\mathbf{k}]})^{\text{old}}. \tag{2.17}$$

.

In contrast to ICA, most IVA algorithms that have been developed to date are based on parametric methods for the estimation of the score function during the adaptation. Originally IVA was formulated for solving the convolutive ICA problem in the frequency domain using multiple frequency bins [47]. This led to the development of IVA-Laplacian (IVA-L) [48, 49], an algorithm that takes higher-order statistics (HOS) into account and assumes a Laplacian distribution for the underlying source component vectors. Conversely, IVA-Gaussian (IVA-G) [6, 94] exploits linear dependencies but does not take HOS into account. However, assuming a Gaussian distribution for the underlying sources simplifies the gradient of the MI objective function and makes the Hessian positive definite. This enables second-order algorithms such as Newton-variants to become practical and promises to improve the quality of convergence. Finally, IVA-generalized Gauussian (IVA-GGD) [8] is a more general IVA implementation where both second and higher order statistics are taken into account. This algorithm assumes a multivariate generalized Gaussian distribution (MGGD) for the underlying sources and through the estimation of its parameters, multivariate Gaussian and Laplacian distributions become special cases.

## 2.4 Nonorthogonal ICA and IVA

Optimization problems with matrix parameters arise in many BSS algorithms. In particular, for the ICA/IVA update rules, performing the optimization procedure on the space of all invertible matrices, may result in poor convergence due to inversion of $\mathbf{W}$ matrix at each iteration. As we mentioned in (2.2.2) and (2.3.2), a potential solution for this issue, is to post multiply the gradient of the objective function by $\mathbf{W}^\top \mathbf{W}$ or $(\mathbf{W}^{[k]})^\top (\mathbf{W}^{[k]})$ respectively. Although this "natural gradient" approach has shown significant results in terms of its convergence properties [4, 24, 27], there are still limitations associated with optimization using matrix parameters. For instance, the term $E\{\boldsymbol{\phi}\mathbf{y}^\top\}$ in (2.10), may be complicated especially when the class of estimated PDFs for the latent sources is broad. This motivates the division of the minimization of (2.4) into a series of subproblems such that we minimize the MI objective function w.r.t. each of the row vectors $\mathbf{w}_1, \ldots, \mathbf{w}_N$ individually. This simplifies the density matching problem as the estimation of a given source will not affect the estimation of the others. In addition, optimization of cost functions w.r.t. each row vector of the demixing matrix provides the following benefits:

- Enables integration of flexible PDF estimation techniques;

- Improvement of the convergence characteristics of the algorithm;

- Simplifies the incorporation of constraints in the ICA/IVA framework;

- Enables the implementation of parallel ICA/IVA algorithms.

A simple approach introduced in [42], is to assume that $\mathbf{W}$ is orthogonal, *i.e.,* $\mathbf{W}\mathbf{W}^\top = \mathbf{I}$. This assumption yields $|\det(\mathbf{W})| = 1$, and allows for the optimization of (2.4) w.r.t. each of the rows of $\mathbf{W}$. Therefore (2.4) reduces to

$$J_{ICA}(\mathbf{W}) = \sum_{n=1}^{N} H(y_n) - H(\mathbf{x}), \tag{2.18}$$

which makes the minimization of (2.4) equivalent to maximization of the negentropy (MN). The natural cost for MN is neg-entropy, which measures the entropic distance of a pdf from that of a Gaussian. Although assuming $\mathbf{W}$ to be orthogonal simplifies the objective function and may improve the stability of the algorithm, the solution space is limited, which may significantly affect the overall separation performance. To avoid this issue we present a decoupling approach that transforms the matrix optimization into a series of vector optimization problems without constraining $\mathbf{W}$ to be orthogonal.

To keep the notation simple, we consider in the following discussion the ICA MI cost function and update rule. An extension to the IVA case can be performed in a straightforward manner. The goal in the decoupling procedure is to decouple the estimation of each row in $\mathbf{W}$, $\mathbf{w}_n^\top$, $n = 1, \ldots, N$. Thus, let $\mathbf{W}_n = [\mathbf{w}_1, \ldots, \mathbf{w}_{n-1}, \mathbf{w}_{n+1}, \ldots, \mathbf{w}_N]^\top \in \mathbb{R}^{(N-1) \times N}$ denote the matrix that contains all rows of $\mathbf{W}$ except the $n$th one. Since the determinant of a matrix is invariant under row permutation up to a sign ambiguity, the square of the det($\mathbf{W}$) term in (2.4) is written as

$$\begin{aligned}
\det(\mathbf{W})^2 &= \det(\mathbf{W}\mathbf{W}^\top) \\
&= \det\left(\begin{bmatrix} \mathbf{W}_n \\ \mathbf{w}_n^\top \end{bmatrix} [\mathbf{W}_n \mathbf{w}_n^\top]\right) \\
&= \det\left(\begin{bmatrix} \mathbf{W}_n \mathbf{W}_n^\top & \mathbf{W}_n \mathbf{w}_n \\ \mathbf{w}_n^\top \mathbf{W}_n^\top & \mathbf{w}_n^\top \mathbf{w}_n \end{bmatrix}\right) \\
&= \det(\mathbf{W}_n \mathbf{W}_n^\top) \mathbf{w}_n^\top (\mathbf{I} - \mathbf{W}_n^\top (\mathbf{W}_n \mathbf{W}_n^\top)^{-1} \mathbf{W}_n) \mathbf{w}_n,
\end{aligned} \quad (2.19)$$

where the term $\mathbf{H}_n = \mathbf{I} - \mathbf{W}_n^\top (\mathbf{W}_n \mathbf{W}_n^\top)^{-1} \mathbf{W}_n$ is the orthogonal projection onto the null space of $\mathbf{W}_n$. By definition, the matrix $\mathbf{H}_n$ is rank one and thus $\mathbf{H}_n = \mathbf{h}_n \mathbf{h}_n^\top$, where $\mathbf{h}_n$ is perpendicular to all row vectors of $\mathbf{W}_n$. Thus,

$$
\begin{aligned}
|\det(\mathbf{W})| &= \sqrt{\det(\mathbf{W}_n \mathbf{W}_n^\top)^2 \mathbf{w}_n^\top \mathbf{h}_n \mathbf{h}_n^\top \mathbf{w}_n} \\
&= \sqrt{\det(\mathbf{W}_n \mathbf{W}_n^\top)^2 (\mathbf{h}_n^\top \mathbf{w}_n)^2} \\
&= |\det(\mathbf{W}_n \mathbf{W}_n^\top)| |(\mathbf{h}_n^\top \mathbf{w}_n)|.
\end{aligned}
\tag{2.20}
$$

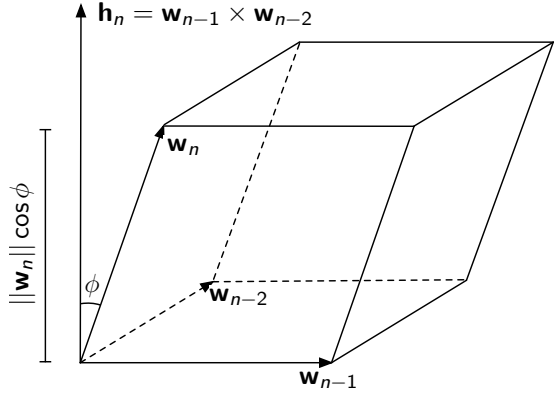

FIG. 2.2. Illustration of the decoupling trick for three vectors.

Therefore, (2.4) can be written as

$$
J_{ICA}(\mathbf{W}) = \sum_{n=1}^{N} H(y_n) - \log |(\mathbf{h}_n^\top \mathbf{w}_n)| - \log |\det(\mathbf{W}_n \mathbf{W}_n^\top)| - H(\mathbf{x}),
\tag{2.21}
$$

where the terms $H(\mathbf{x})$ and $\log |\det(\mathbf{W}_n \mathbf{W}_n^\top)|$ are independent of $\mathbf{w}_n$. The gradient of (2.21) w.r.t. $\mathbf{w}_n$ is given by

$$
\frac{\partial J_{ICA}(\mathbf{W})}{\partial \mathbf{w}_n} = -E\{\phi(y_n)\mathbf{x}\} - \frac{\mathbf{h}_n}{\mathbf{h}_n^\top \mathbf{w}_n},
\tag{2.22}
$$

Therefore, the estimate of $\mathbf{W}$ can be determined w.r.t. each row vector $\mathbf{w}_n$, $n = 1, \ldots, N$ independently, by using the gradient update rule

$$
(\mathbf{w}_n)^{\text{new}} \leftarrow (\mathbf{w}_n)^{\text{old}} - \gamma \frac{\partial J_{ICA}(\mathbf{W})}{\partial \mathbf{w}_n}.
\tag{2.23}
$$

A simple three-dimensional geometric interpretation of the decoupling trick is illus-trated by Fig. (2.2). In this three dimensional case, the term $|\det(\mathbf{W})|$ represents the volume of a parallelepiped spanned by the vectors $\mathbf{w}_n, \mathbf{w}_{n-1}, \mathbf{w}_{n-2}$, and is written as the product of the area of its base and its height.

## 2.5 Summary

In this chapter, we present the mathematical formulation of the BSS problem and discuss the role of diversity in BSS. We present the ICA and IVA models and derive their ML objective functions. Then, we show their relation to the MI objective functions as the number of samples approaches infinity. Finally, we demonstrate the benefits of the nonorthogonal ICA and IVA framework and use this framework to develop effective ICA algorithms, discussed more in Chapters 3,5, as well as IVA algorithms, discussed more in Chapter 4. These algorithms we show to be effective in a variety of applications.

# Chapter 3

# A NEW FLEXIBLE ICA ALGORITHM

In order to achieve the optimality conditions for ICA using the ML framework, exact knowledge of the true PDF of the latent sources is required. However, this information is usually not available in most real-world applications. Therefore, accurate estimation of source PDFs is critical in order to avoid model mismatch and therefore poor ICA performance. In this chapter, we present a new and efficient ICA algorithm based on entropy maximization with kernels, (ICA-EMK), which uses both global and local measuring functions as constraints to dynamically estimate the PDF of the sources with reasonable complexity. In addition, the new algorithm performs optimization with respect to each of the cost function gradient directions separately, enabling parallel implementations on multi-core computers. We demonstrate the superior performance of ICA-EMK over competing ICA algorithms using simulated as well as real-world data.

## 3.1 Maximum Entropy Principle

The estimation of a PDF for given data is a common problem in a wide variety of fields ranging from physics to statistics. Within the field of machine learning, many estimation, detection, and classification problems require knowledge of the data's PDF either explicitly or implicitly, see *e.g.*, [1]. Thus, effective characterization of the density is vital to the success of these machine learning approaches.

Classical density estimation techniques can be characterized as either parametric or nonparametric. The non-parametric methods, such as histogram estimation, k-nearest neighbors (kNN), and kernel density estimation (KDE) [17,46], can provide flexible density matching. Although non-parametric techniques are not limited to any specific distribution, they are generally computationally demanding, especially when sample size is large, and they highly depend on the choice of tuning parameters. For instance, histogram, KNN, and KDE, highly depend on the choice of parameters, such as the number of bins, number of samples in a neighborhood, and the bandwidth for histogram, respectively. Many of the other methods assume a parametric model, such as the GGD for the density. Parametric methods provide a simple form for the PDF and are computationally efficient, however they are limited when the underlying distribution of the data deviates from the assumed parametric form. For example, the GGD limits the PDFs to these that are symmetric and unimodal, and controlling the shape of the distribution through a single parameter.

Semi-parametric methods combine the flexibility of the non-parametric techniques with the relatively simple density form of the parametric techniques. Multidimensional Gaussian mixture model (MGMM) has been widely used for semi-parametric density estimation [17]. However, MGMM is typically time consuming and since the kernel function is limited to only a single type, the trade off between flexibility and generalization ability needs to be carefully weighted when selecting the number of mixtures and their parameters. In contrast, semi-parametric methods, such as those based on the maximum entropy principle [10, 34, 57], combine the simple density form and the flexibility of nonparametric and parametric methods, yielding a unique solution provided by the maximum entropy principle.

The maximum entropy principle, can be described by the optimization problem [31]

$$\begin{aligned}\max_{p(x)} \quad & H(x) = -\int_{-\infty}^{\infty} p(x) \log p(x) \, dx \\ \text{s.t.} \quad & \int_{-\infty}^{\infty} r_i(x) p(x) \, dx = \alpha_i, \text{ for } i = 1, \ldots, M,\end{aligned} \quad (3.1)$$

where $r_i(x)$ are the measuring functions such as, $\alpha_i = \sum_{t=1}^{T} r_i(t)/T$ are the sample averages, and $M$ denotes the total number of measuring functions. To ensure that $p(x)$ is a valid PDF, we select $\alpha_1 = r_1(x) = 1$. The constrained optimization problem (3.1) can be written as an unconstrained one through the Lagrangian function:

$$L(p(x)) = H(p(x)) + \sum_{i=1}^{M} \lambda_i \int_{-\infty}^{\infty} (r_i(x) - \alpha_i) p(x) \, dx, \quad (3.2)$$

where $\lambda_i$, $i = 1, \ldots, M$ are the Lagrange multipliers. By differentiating (3.2) with respect to $p(x)$ and setting its derivative equal to zero, the equation of the maximum entropy distribution is obtained as

$$\hat{p}(x) = \exp\left\{-1 + \sum_{i=1}^{M} \lambda_i r_i(x)\right\}, \quad (3.3)$$

and the Lagrange multipliers can be numerically determined to satisfy the constraints in (3.1).

### 3.1.1 Entropy Maximization with Kernels

Entropy maximization with kernels (EMK) is a robust semiparametric method that has been shown to provide desirable estimation performance [34]. Its flexibility to model a wide range of distributions and its simple mathematical form make it a particularly attractive candidate for ICA. The performance of EMK is achieved by using both global as well as adaptive local measuring functions to provide constraints on the overall statistics and gain insight into the local behavior of the source PDFs, respectively.

One of the main components of EMK is the numerical estimation of the Lagrange multipliers given in (3.3). For a given set of measuring functions, the Lagrange multipliers, are estimated using the Newton iteration method [34]

$$\lambda^{(k+1)} = \lambda^{(k)} - \mathbf{J}^{-1} E_{p(x)^{(k)}} \{\mathbf{r} - \alpha\}, \tag{3.4}$$

where $p(x)^{(k)}$ it the estimated PDF for the $k$th iteration, and $\mathbf{r}$, $\lambda$, $\alpha$ denote the $M$-dimensional vector of measuring functions, Lagrange multipliers, and sample averages, respectively. The $(i, j)$th entry of the Jacobian matrix $\mathbf{J}$ is given by

$$\int_{-\infty}^{\infty} r_i(x) r_j(x) p^{(k)}(x) dx, \tag{3.5}$$

and the $i$th entry of $E_{p^{(k)}} \{\mathbf{r} - \alpha\}$ is given by

$$\int_{-\infty}^{\infty} (r_i(x) - \alpha_i) p^{(k)}(x) dx. \tag{3.6}$$

We select the global measuring functions $\{1, x, x^2, x/(1 + x^2)\}$ to relate to sample estimates of the PDF, mean, variance, and HOS, respectively. For local measuring functions, we use a number of Gaussian kernels given by the set $\{e^{(-(x-\mu_i)^2/2\sigma_i^2)}\}$, $i = M - 4$. The number of local measuring functions is chosen by an information-theoretic criterion, the minimum description length (MDL) [78, 95]. For each Gaussian kernel the parameters $\mu$ and $\sigma^2$ are estimated by finding the greatest deviation between the estimated and the true PDF. For further details about the choice of the local measuring functions and the estimation of their parameters, we refer the reader to [34]. In what follows we present a new ICA algorithm, ICA-EMK, that takes advantage of the accurate yet analytically simple estimation capability of EMK and yields an algorithm with superior separation performance to competitive ICA algorithms.

## 3.2 ICA-EMK

Using the decoupled MI objective function as the starting point we can see that the differential entropy of the $n$th source estimate is a function of its unknown PDF. Therefore, by using the Lagrange multiplier estimates from (3.4), the differential entropy of the $n$th source estimate can be written as

$$H(y_n) = -E\left\{-1 + \sum_{i=1}^{M} \lambda_i r_i(y_n)\right\} = 1 - \sum_{i=1}^{M} \lambda_i \alpha_i.$$

This allows us to rewrite (2.4) as a sequence of cost functions given by

$$J_{ICA}(\mathbf{w}_n) = 1 - \sum_{i=1}^{M} \lambda_i(n)\alpha_i(n) - \log\left|\mathbf{h}_n^\top \mathbf{w}_n\right| - C, \tag{3.7}$$

where $\lambda_i(n)$ and $\alpha_i(n)$ denote the estimated Lagrange multipliers and sample averages for each of the source estimates. The gradient of (3.7) w.r.t. $\mathbf{w}_n$ is given by

$$\frac{\partial J_{ICA}(\mathbf{w}_n)}{\partial \mathbf{w}_n} = -\sum_{i=1}^{M} \lambda_i E\left\{\frac{\partial r_i(y_n)}{\partial y_n}\mathbf{x}\right\} - \frac{\mathbf{h}_n}{\mathbf{h}_n^\top \mathbf{w}_n}. \tag{3.8}$$

Performing the optimization routine in a Riemannian manifold rather than a classical Euclidean space provides important convergence advantages. Therefore, following [57], we define the domain of our cost function to be the unit sphere in $\mathbb{R}^N$. Then, by using the projection transformation onto the tangent hyperplane of the unit sphere at the point $\mathbf{w}_n$, the normalized gradient of our cost function is given by

$$\mathbf{u}_n = \mathbf{P}_n(\mathbf{w}_n)\frac{\partial J_{ICA}(\mathbf{w}_n)}{\partial \mathbf{w}_n}, \tag{3.9}$$

where $\mathbf{P}_n(\mathbf{w}_n) = \mathbf{I} - \mathbf{w}_n \mathbf{w}_n^T$ and $\|\mathbf{w}_n\| = 1$. A pseudo-code description of the ICA-EMK algorithm is given in Algorithm 1 below. The main part of this algorithm is the loop described in lines 3–10. The algorithm terminates when $|J_{ICA}(\mathbf{W}_{iter}) - J_{ICA}(\mathbf{W}_{iter-k})| < \delta$, where $\delta$ is a tolerance chosen by the user, $k$ is a small integer that desensitizes the algorithm to fluc-

tuations of the cost function $J_{ICA}(\mathbf{W})$. The loop also terminates if the number of iterations exceeds a pre-defined maximum number of iterations.

---
**Algorithm 1** ICA-EMK
---
1: **Input**: $\mathbf{X} \in \mathbb{R}^{N \times T}$
2: Initialize $\mathbf{W}_0 \in \mathbb{R}^{N \times N}$
3: **for** $n = 1:N$ **do**
4:     Given $\{r_i\}_{i=1}^{M}$, estimate Lagrange multipliers using (3.4)
5:     Compute $\mathbf{h}_n$, orthogonal to $\mathbf{w}_i$ for all $i \neq n$
6:     Calculate the derivative $\frac{\partial J_{ICA}(\mathbf{w}_n)}{\partial \mathbf{w}_n}$ using (3.8)
7:     Project the gradient onto the unit sphere using (3.9)
8:     $(\mathbf{w}_n)^{\text{new}} \leftarrow (\mathbf{w}_n)^{\text{old}} - \gamma \mathbf{u}_n$
9: **end**
10: Repeat steps 3 through 10 until convergence in $J_{ICA}(\mathbf{W})$ or until the maximum number of iterations is exceeded
11: **Output**: $\mathbf{W}$
---

### 3.2.1 Parallel Implementation and Performance

In many applications encountered in practice, the number of sources can be quite large subjecting traditional sequential source separation algorithms to lengthy execution times. Since the bulk of the computational complexity of ICA-EMK occurs in lines 3–10 in Algorithm 1, distributing separate iterations of the main loop to separate computation resources is desirable to reduce the total execution time.

The performance improvement to be gained from using a faster mode of execution is limited by the fraction of the time the faster mode is used. This is known as Amdahl's Law and is given by [39]:

$$\text{Speedup} = \frac{t_{\text{old}}}{t_{\text{new}}} = \frac{1}{(1-f) + \frac{f}{s}}, \qquad (3.10)$$

where $t_{\text{old}}$ is the execution time prior to the enhancement, $t_{\text{new}}$ is the execution time after the enhancement, $f \leq 1$ is the fraction of $t_{\text{old}}$ spent on the code to be enhanced, and $s \geq 1$ is the speedup of the enhanced code. In repeated experimental runs of ICA-EMK, $f$ was found to be quite high, on the order of $f > 0.95$ leading to a speedup on the order of $s$.

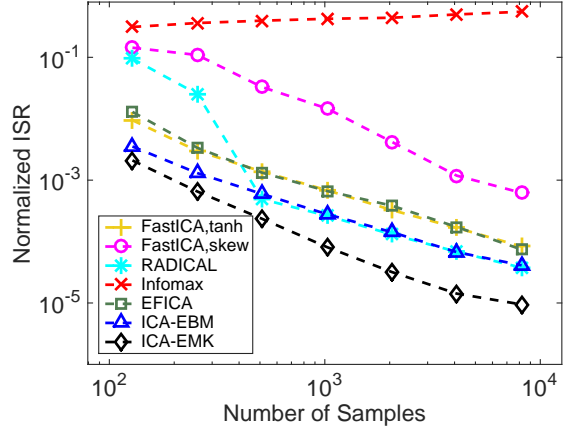

FIG. 3.1. Performance comparison of seven ICA algorithms in terms of the normalized average ISR as a function of the number of samples. The $N = 8$ sources are mixtures of GGDs. Each point is the result of 100 independent runs.

The decoupling trick provides independence between the computation of each of the cost function gradient directions enabling direct exploitation of the natural parallelism on multi-processor or multi-core computers. This is performed by outsourcing the computation of each gradient direction (3.8) to a separate processor or core subject to availability of computing resources. The results from the separate cores are joined in each iteration to evaluate the termination criterion. The overhead associated with forking, then joining, execution streams leads to $s$ being less than, yet very close to, $L$, the number of cores or processors available for parallel execution. Despite the noted overhead, real world applications with a sufficiently large number of sources and samples do achieve significant speedup as our experimental results in Section 3.3.3 demonstrate.

### 3.3 Experimental Results

We demonstrate the performance of ICA-EMK, in terms of its separation power, using simulated as well as natural images as sources. We compare ICA-EMK with six commonly used ICA algorithms: FastICA, using the symmetric decorrelation approach with two nonlinearities tanh and skew (FastICAtanh) and (FastICAskew), RADICAL, Infomax, EFICA, and ICA-EBM. FastICAtanh favors symmetric distributions and FastICAskew skewed ones. RADICAL is a nonparametric algorithm that can successfully accommodate more complex PDFs. Infomax is based on a fixed super-Gaussian source model. EFICA is an efficient FastICA version that uses the univariate generalized Gaussian distribution (GGD) source model. ICA-EBM favors distributions that are skewed, heavy or light-tailed and bimodal. Moreover, we also quantify the performance of ICA-EMK, in terms of execution time, using simulated data. We elect to limit the maximum number of local measuring functions to 5 so as to control complexity. We observed that the overall impact of this limitation in terms of performance is negligible. In each of the following experiments, ICA-EMK is initialized using a random matrix $\mathbf{W}$.

#### 3.3.1 Simulated Data

In the first experiment, we generate 8 simulated sources each of which is a mixture of GGD kernels. The PDF of each source is given by [72]

$$p(x; \beta_i, \mu_i, \sigma_i) = \sum_{i=1}^{K} \pi_i \eta_i \exp\left(-\frac{(x - \mu_i)^{2\beta_i}}{2\sigma_i^{2\beta_i}}\right), \ x \in \mathbb{R}$$

where $\eta = \frac{\beta}{2^{\frac{1}{2\beta}} \Gamma(\frac{1}{2\beta}) \sigma}$ and $K$ is the number of mixtures. To generate sufficiently complicated sources, $K$ is randomly selected to be either 4 or 5. The weight parameters $\pi_i$ are randomly selected from the interval $(0, 1)$ such that $\sum_{i=1}^{K} \pi_i = 1$. The shape parameter $\beta$ is randomly selected from the interval $(0.25, 4)$. Note that $\beta$ controls the peakedness and spread of the distribution. If $\beta < 1$, the distribution is more peaky than Gaussian with heavier tails, and if $\beta > 1$, it is less peaky with lighter tails. When $K = 4$, the GGD means are chosen to be $\{-8, -4, 4, 8\}$, whereas when $K = 5$ the means are chosen to be $\{-10, -5, 0, 5, 10\}$. In the second experiment, we generate 8 sources using the Gamma distribution with PDF $p(x) = x^{\beta-1} \exp(-x)$, $x \geq 0$. For each of the sources the shape parameter takes values from the set $\{1, 2, \ldots, 8\}$, resulting in different unimodal skewed PDFs. For both experiments, the sources are mixed by a random square matrix whose elements are drawn from a zero mean, unit variance Gaussian distribution. To evaluate the performance of our algorithm we use the average interference to signal ratio (ISR) as in [57]. The rest of the algorithm parameters are $k = 8$, and $M = 9$. Results are averaged over 100 independent runs.

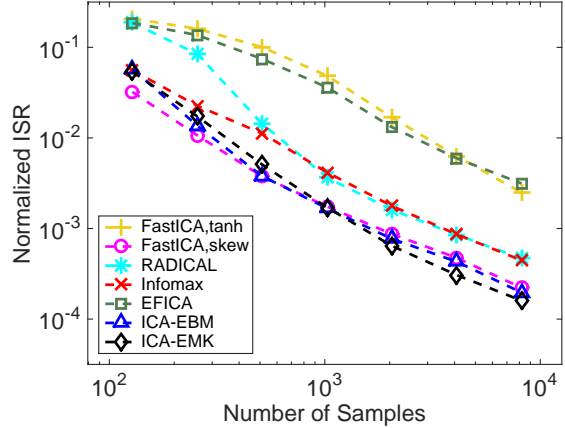

FIG. 3.2. Performance comparison of seven ICA algorithms in terms of the normalized average ISR as a function of the number of samples. The $N = 8$ sources are drawn from the Gamma distribution with different shape parameters. Each point is the result of 100 independent runs.

In Fig. 3.1 we observe that ICA-EBM and RADICAL exhibit good performance as the sample size increases revealing the flexibility of their underlying density models. On the other hand, the two different versions of FastICA, EFICA, and Infomax do not perform well due to their simple underlying density model. Overall, ICA-EMK performs the best

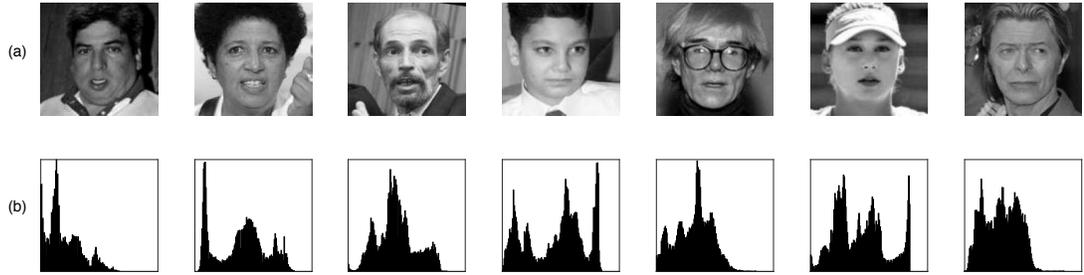

FIG. 3.3. Seven face images with complicated densities. (a) Original grayscale sources images of size $168 \times 168$, (b) Histogram of each image where the number of bins is 128.

among the seven algorithms.

In Fig. 3.2, we see that FastICAskew performs the best when the sample size is less than 1000. When the sample size becomes greater than 1000, the performance of ICA-EBM is similar to that of FastICAskew since the large sample size allows for accurate approximation of the differential entropy of the estimated sources. For smaller sample sizes and simpler distributions, RADICAL does not perform well. When the sample size becomes greater than 1000, its performance is very similar to Infomax's performance. FastICAtanh and EFICA do not provide good performance compared with the other algorithms due to the inherent model mismatch. Finally, ICA-EMK for large sample sizes provides the best performance, since the probability density model is most accurately estimated at each ICA iteration.

Despite its superior separation performance, ICA-EMK is computationally demanding compared with other algorithms—with the exception of RADICAL, which is the most costly for large sample sizes. The additional time penalty incurred diminishes, however, with the increase in parallelism of the computing resource at hand. The trade-off between superior performance and diminishing time penalty, hence does favor the use of ICA-EMK over others for improved separation performance.

### 3.3.2 Mixture of Atificial Images

To show the improvement that ICA-EMK provides over FastICA and ICA-EBM, in this experiment we use seven face images as independent sources. Fig. 3.3 shows the grayscale images obtained from [40, 97] as well as their associated histograms. It is clear from the histograms that the images represent a wide range of complicated source distributions.

To setup the experiment, we create the independent sources by vectorizing the $168 \times 168$ images and then linearly mix them using a random mixing matrix. After obtaining the estimated demixing matrices from each of the algorithms, we estimate the independent components and, together with their associated demixing vectors, pair them with the true sources. In the case where more than one estimated component is paired with a single true source, we use Bertsekas algorithm [13] to find the best assignment as described in [81]. To evaluate the performance of the three algorithms, we use the absolute value of the correlation between the true and the estimated sources. Results are averaged over 300 independent runs. In Fig. 3.4, we observe that ICA-EBM performs significantly better than FastICA for all but two images. However, ICA-EMK provides the best performance among the three algorithms overall.

### 3.3.3 Parallel Implementation Performance

To demonstrate the computational speedup of the parallel ICA implementation over its sequential counterpart (ICA-EMK where the decoupled source computations are forced to run on a single processing core), we compare the average execution time of each implementation on the 5 GGD mixtures of simulated sources from the prior subsection with $T = 1000$ samples. Both implementations are performed in the Matlab environment on a lab computer with a quad-core processor and 8 GB of RAM. Figure 3.5 shows the result of running this experiment for a number of sources $N \in \{2, 4, 8, 16, 32, 64, 128\}$ where each point is the result of the average of 100 independent runs. Both algorithms execute 100 iterations irrespective of convergence properties. The red and blue curves, associated with the y-axis to the left, represent the average CPU time for the non-parallel ICA and parallel ICA implementations respectively. The green curve, associated with the y-axis on the right, represents the speedup as a result of exploiting parallelism. We observe that when the number of sources $N = 2$, the speedup is small, since two of the four cores are idling. As the number of sources increases, the speedup improves and approaches the number of processor cores without reaching it—inline with our discussion following equation (3.10). This is due to the overhead associated with forking then joining the computation in addition to the fact that the ICA algorithm is not fully parallelizable and some sequential portions remain. Similarly, with $L$ CPUs, a speedup slightly below of $L$ can be expected as long as there is a sufficiently large number of sources to keep processor utilization near 100%.

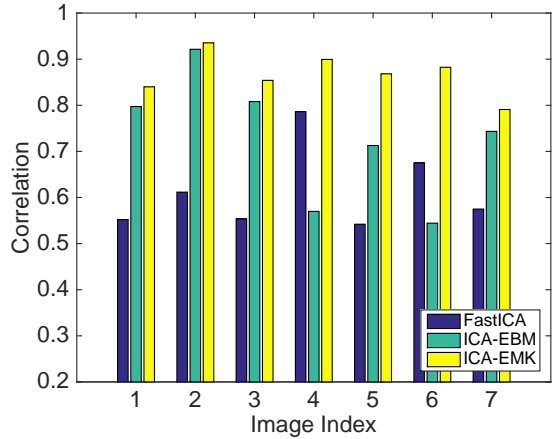

Fig. 3.4. Correlation between the true and estimated source images using FastICA (blue bars), ICA-EBM (green bars), and ICA-EMK (yellow bars) algorithms. Results are averaged over 300 independent runs.

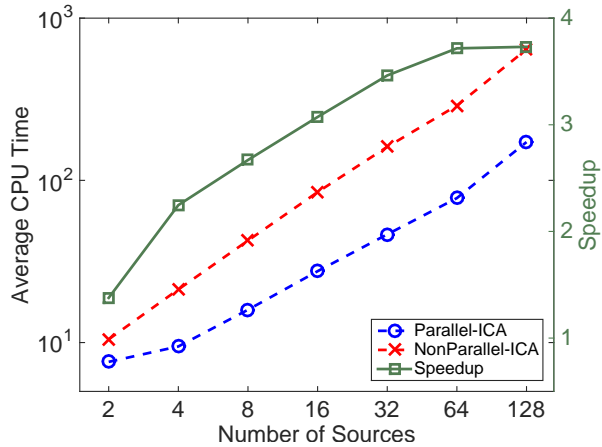

FIG. 3.5. Average CPU time for parallel and sequential implementations of ICA-EMK as a function of the number of sources and the resulting speedup.

## 3.4 Summary

In this chapter, we present a new and efficient ICA algorithm, ICA by entropy maximization with kernels, that uses both global and local measuring functions to provide accurate estimates of the PDFs of the source estimates. ICA-EMK has been implemented in a parallel fashion so that it is computationally attractive when the number of sources and number of cores increase. Experimental results confirm the attractiveness of the new ICA algorithm that can separate sources from a wide range of distributions. We also provide a mathematical justification of the convergence of the ICA-EMK algorithm. We should note that, due to its flexibility, ICA-EMK can be used in many applications especially when prior knowledge about the data is not available. Where prior knowledge of the PDF exists, however, the estimation technique can be adjusted based upon the needs of the application.

# Chapter 4

# A CLASS OF EFFECTIVE IVA ALGORITHMS BASED ON MGGD SCV MODEL

IVA is a recent extension of ICA to multiple datasets that makes full use of the statistical dependence across multiple datasets to achieve source separation. In Chapter 2, we showed how IVA can be formulated in a ML framework. Despite the potential usefulness of IVA in a variety of applications, the density models proposed to date are limited in flexibility, primarily due to the fact that efficient modeling of multivariate PDFs is a more challenging task. This motivates the development of effective models for the underlying source density and appropriate methods for the estimation of the parameters for the models.

The MGGD provides an effective model for the latent multivariate sources in IVA, however, the performance of IVA algorithms that utilize this density models highly depends on the estimation of the MGGD distribution parameters—namely, the scatter matrix and the shape parameter. In this chapter, we discuss techniques to estimate the parameters of the MGGD and successfully integrate them into IVA. Current methods that attempt to estimate the scatter matrix for a given value of the shape parameter, provide undesirable performance when the value of the shape parameter increases. We begin this chapter by presenting two different ML techniques to overcome the issue of estimating the scatter matrix for large values of the shape parameter. The first is based on a Fisher scoring (FS) algorithm and the second on a fixed point (FP) algorithm. Using these two estimation techniques, we derive a class of IVA algorithms that estimate the shape parameter and scatter matrix jointly, while taking both SOS and HOS into account. We conclude the chapter with a series of simulations, demonstrating the significant advantages that are offered by these adaptive IVA algorithms over other competing algorithms.

## 4.1 MGGD and its Parameters

The probability density function of an MGGD is given by [51]

$$p(\mathbf{y}; \mathbf{\Sigma}, \beta, m) = \frac{\Gamma\left(\frac{K}{2}\right)}{\pi^{\frac{K}{2}} \Gamma\left(\frac{K}{2\beta}\right) 2^{\frac{K}{2\beta}}} \frac{\beta}{m^{\frac{K}{2}} |\mathbf{\Sigma}|^{\frac{1}{2}}} \exp\left[-\frac{1}{2m^\beta} \left(\mathbf{y}^\top \mathbf{\Sigma}^{-1} \mathbf{y}\right)^\beta\right], \quad (4.1)$$

where $K$ is the dimension of the probability space, $\mathbf{y} \in \mathbb{R}^K$ is a random vector, $m$ is the scale parameter, $\beta > 0$ is the shape parameter that controls the peakedness and the spread of the distribution and $\mathbf{\Sigma}$ is a $K \times K$ symmetric positive scatter matrix. If $\beta = 1$, the MGGD is equivalent to the multivariate Gaussian and the matrix $\mathbf{\Sigma}$ becomes the covariance matrix. If $\beta < 1$ the distribution of the marginals is more peaky and has heavier tails, while $\beta > 1$ is less peaky and has lighter tails.

Recently, the estimation of the parameters of MGGD has received significant attention, due to the fact that MGGD has numerous applications including those in video coding, image denoising, and medical image analysis [3, 29, 32, 53, 98]. Existing approaches to this problem, see *e.g.*, [19, 74, 76, 86, 92, 93, 99], attempt to estimate $\mathbf{\Sigma}$ for a given value of $\beta$. However, their accuracy suffers when the value of $\beta$ becomes large, which makes these techniques unsuitable for many applications.

Existing approaches to this problem, see [93],[92] and [19, 74, 76, 86, 99], attempt to estimate the scatter matrix for a given value of the shape parameter. However, their accuracy suffers when the value of shape parameter becomes large, which makes them unsuitable for many applications. In [93],[92], method of moments (MoM) and ML techniques, were explored. In [76], authors prove that the scatter matrix ML estimator exists and is unique for any value of the shape parameter that belongs to the interval (0, 1). Although simulation results reveal the unbiasedness and consistency of the ML estimators of the scatter matrix and the shape parameter of the distribution, numerical results show that they provide highly inaccurate results when $\beta \geq 2$.

### 4.1.1 ML-FS

Let $(\mathbf{y}_1, \mathbf{y}_2, ..., \mathbf{y}_T)$ be a random sample of $T$ observation vectors of dimension $K$, drawn from a zero mean MGGD with parameters $\beta$ and $\Sigma$, and $m$. The corresponding ML estimates $\hat{\beta}$, $\hat{\Sigma}$ and $\hat{m}$ are found by solving the ML equations, described in the following. Assume first $\beta$ is known. The ML estimate $\hat{\Sigma}$ is the solution of the following equation [76]

$$\Sigma = \sum_{i=1}^{N} \frac{p}{u_i + u_i^{1-\beta} \sum_{i \neq j} u_j^{\beta}} \mathbf{y}_i \mathbf{y}_i^{\top}, \qquad (4.2)$$

for unknown $\Sigma$, where $u_i = \mathbf{y}_i^{\top} \Sigma^{-1} \mathbf{y}_i$. Once $\hat{\Sigma}$ has been computed, $\hat{m}$ is immediately given by

$$\hat{m} = \left( \frac{1}{N} \sum_{i=1}^{N} \hat{u}_i^{\beta} \right)^{\frac{1}{\beta}}, \qquad (4.3)$$

where $\hat{u}_i = \mathbf{y}_i^{\top} \hat{\Sigma}^{-1} \mathbf{y}_i$.

In the general case where $\beta$ is unknown, $\hat{\Sigma}$ and $\hat{\beta}$ are found by solving equation (4.2), along with

$$\gamma(\beta) = \frac{pN}{2 \sum_{i=1}^{N} u_i^{\beta}} \sum_{i=1}^{N} \left[ u_i^{\beta} \ln(u_i) \right] - \frac{pN}{2\beta} \left[ \Psi\left(\frac{p}{2\beta}\right) + \ln 2 \right] - N - \frac{pN}{2\beta} \ln\left( \frac{\beta}{pN} \sum_{i=1}^{N} u_i^{\beta} \right) = 0, \quad (4.4)$$

whose solution is $\hat{\beta}$. Here, $\Psi$ is the digamma function. Once the solutions $\hat{\Sigma}$ and $\hat{\beta}$ of (4.2) and (4.4) have been found, $\hat{m}$ is computed directly from (4.3).

It is seen from the above that the main difficulty, in the computation of $\hat{\beta}$, $\hat{\Sigma}$ and $\hat{m}$, lies in solving equation (4.2). There is no closed form solution for the ML estimation of these parameters. Hence, we use an iterative scheme to simultaneously estimate $\Sigma$ and $\beta$ using (4.2) and (4.4) respectively.

The likelihood function of $(\mathbf{y}_1, \mathbf{y}_2, ..., \mathbf{y}_T)$ where $m$ has been replaced by its estimate in (4.3) is given by

$$\mathcal{L}(\Sigma; \Theta) = \prod_{i=1}^{T} p(\mathbf{y}_i; \Theta) = \left[ \frac{\beta \Gamma\left(\frac{K}{2}\right)}{\pi^{\frac{K}{2}} \Gamma\left(\frac{K}{2\beta}\right) 2^{\frac{K}{2\beta}}} \right]^T \left( \frac{KT}{\beta} \right)^{\frac{KT}{2\beta}} \exp\left( -\frac{KT}{2\beta} \right) \times \left[ \frac{1}{|\Sigma|} \left( \sum_{i=1}^{T} u_i^\beta \right)^{-\frac{K}{\beta}} \right]^{\frac{T}{2}}, \quad (4.5)$$

where $\Theta$ denotes the parameter space that contains the entries of the scatter matrix $\Sigma$, $m$ and $\beta$. By fixing $\beta$, the likelihood function depends only on the entries of $\Sigma$. Defining the gradient of the likelihood function similar to [76], we introduce the functional

$$F : \mathcal{S}_+^K \to \mathbb{R}^+ \setminus \{0\}$$

$$\Sigma \mapsto |\Sigma|^{-1} \left( \sum_{i=1}^{T} u_i^\beta \right)^{-\frac{K}{\beta}},$$

where $\mathcal{S}_+^K$ is the space of all real $K \times K$ symmetric and positive definite matrices. By omitting the constant term from (4.5), the gradient of the likelihood can then be written as

$$\nabla \mathcal{L}(\Sigma; \Theta) = [F(\Sigma)]^{\frac{T-2}{2}} \nabla F(\Sigma),$$

and the gradient of $F$ at a point $\Sigma \in \mathcal{S}_+^K$ is given by [76]

$$\nabla F(\Sigma) = F(\Sigma) \Sigma^{-1} [f(\Sigma) - \Sigma] \Sigma^{-1},$$

where

$$f : \mathcal{S}_+^K \to \mathcal{S}_+^K$$

$$\Sigma \mapsto \sum_{i=1}^{T} \frac{K}{u_i + u_i^{1-\beta} \sum_{i \neq j} u_j^\beta} \mathbf{y}_i \mathbf{y}_i^\top.$$

To numerically maximize the likelihood function we use a variation of the Newton-Raphson optimization algorithm called the ML-FS algorithm. Since the negative inverse Hessian has been replaced by the inverse of the Fisher information matrix, we need to calculate its entries.

To calculate the entries of the Fisher information matrix, we first define the manifold $\mathcal{M}$ of a zero-mean MGGD as well as an associated metric. The MGGD manifold is parameterized by $\beta$ and the matrix $\Sigma$. In our particular case since $\beta$ is fixed, $\mathcal{M}$ is parameterized only by the entries of $\Sigma \in \mathcal{S}_+^K$, so $\mathcal{M}$ is isomorphic to $\mathbb{R}^n$ where $n = \frac{K(K+1)}{2}$. Each point that lies on $\mathcal{M}$ is a probability density function. To measure the distance between two pdfs we need to calculate the length of the curve that connects those two points and has minimum length. This curve is called a geodesic path and is determined through the elements of the Fisher information matrix. Thus, if $\Theta = (\theta_1, \theta_2, ..., \theta_n)$ denotes the parameter space of $\mathcal{M}$, the Fisher metric is defined by the matrix elements

$$\mathbf{G}_{ij}(\theta) = -E\left\{\frac{\partial^2}{\partial \theta_i \partial \theta_j} \ln p(\mathbf{y}; \Theta)\right\}, \quad i, j = 1, ...n,$$

where $\mathbf{G}$ is an $n \times n$ matrix. Trying to define geodesic paths within an MGGD manifold, [93] proposed a simpler way for the elements of the metric defined on a $K$-dimensional sub-manifold. Thus the entries of the Fisher information matrix are defined by

$$\mathbf{G}_{ii}(\beta) = \frac{1}{4}\left(\frac{3K + 6\beta}{K + 2} - 1\right), \tag{4.6}$$

and

$$\mathbf{G}_{ij}(\beta) = \frac{1}{4}\left(\frac{K + 2\beta}{K + 2} - 1\right), \quad i \neq j, \tag{4.7}$$

for $i, j = 1, ..., K$. From (4.6) and (4.7), we see that Fisher information matrix depends only on the fixed value of $\beta$ and the dimension $K$. This results in the reduction of the computational cost since the update of the inverse of the Fisher information matrix is not required. Fisher scoring iteration is defined as

$$\Sigma^{(k+1)} = \Sigma^{(k)} + \mathbf{G}^{-1} \nabla \mathcal{L}(\Sigma; \Theta). \tag{4.8}$$

MoM provides an effective and efficient initialization for the ML-FS algorithm with only a few steps of the algorithm being sufficient to obtain good estimates.

A pseudo-code description of the ML-FS algorithm is given in Algorithm 2 below. The main part of this algorithm is the loop described in lines 4-11. The algorithm exits this loop when $D(k) <$ tol, where $D(k)$ is the relative difference between two successive estimates, and tol is a tolerance bound, chosen by the user. The loop is also terminated whenever the number of iterations exceeds a pre-defined upper bound $N_{max}$. Although second order

---

**Algorithm 2** ML-FS

1: **Input**: $\mathbf{X} \in \mathbb{R}^{K \times T}$, optionally $\beta$
2: Initialize $\Sigma$ using MoM
3: If $\beta$ is not given initialize both $\Sigma$ and $\beta$ using MoM
4: **while** $(D(k) >$ tol$)$ and $(k < N_{max})$ **do**
5:     Estimate $\Sigma$ using one iteration of (4.8)
6:     **if** $\beta$ is not given **then**
7:         Estimate $\beta$ by applying Newton-Raphson into (4.4)
8:     **else**
9:         Go to step 5
10:     **end**
11: **end**
12: Using $\Sigma$ and $\beta$, estimate $m$ with (4.3)
13: **Output**: $\Sigma, \beta, m$

---

optimization algorithms provide desirable performance in terms of their convergence properties, they highly depend on the initialization. A more general class of algorithms that are invariant of the initial value are those based on the fixed point iteration methods.

### 4.1.2 RA-FP

With regard to ML estimators, it has become clear, from [19, 74, 76, 86, 99], that they can be computed using FP algorithms. As in [76][86], it is possible to formulate (4.2) as a fixed point equation. Considering the function (4.6) we observe that is just the right hand side of equation (4.2). Therefore, this equation can be written

$$\Sigma = f(\Sigma) \tag{4.9}$$

which is indeed a fixed point equation. In other words, the ML estimate $\hat{\Sigma}$ is the solution of the fixed point equation (4.9) associated with the function $f$ defined in (4.6). It is well-known that the solution of a fixed point equation, such as (4.9) may be attempted using an FP algorithm, which gives successive fixed point iterates

$$\Sigma_{k+1} = f(\Sigma_k) \qquad k = 0, 1, 2, \ldots \tag{4.10}$$

Indeed, this algorithm was used in [76][86]. Concretely, it consists in repeating (4.10) until the iterates $\Sigma_k$ stabilize. That is, until there no sensible difference between $\Sigma_k$ and $\Sigma_{k+1}$.

The convergence of the FP algorithm (4.10) depends on the function $f$ being contractive. In the present context, numerical experiments show the function $f$, (which depends on $\beta$ as can be seen in (4.6)), is not contractive when $\beta \geq 2$. The proposed RA-FP algorithm, overcomes this difficulty. It has been shown in [76][86], that the FP algorithm (4.10) gives accurate estimates of $\Sigma$ when $\beta < 2$. The main contribution of the present paper is to describe the new RA-FP algorithm, which is a generalization of the FP algorithm (4.10), and is capable of producing accurate estimates of $\Sigma$ when $\beta \geq 2$.

The RA-FP algorithm uses the Riemannian geometry of the space $\mathcal{S}_+^K$. Precisely, it implements successive Riemannian averages of fixed point iterates. The definition of the Riemannian average of $\mathbf{P}, \mathbf{Q} \in \mathcal{S}_+^p$ is the following. For $t \in [0, 1]$, the Riemannian average with ratio $t$ of $\mathbf{P}$ and $\mathbf{Q}$ is $\mathbf{P}\#_t\mathbf{Q}$, given as in [86]

$$\mathbf{P}\#_t\mathbf{Q} = \mathbf{P}^{1/2}(\mathbf{P}^{-1/2}\mathbf{Q}\mathbf{P}^{-1/2})^t\mathbf{P}^{1/2}, \qquad (4.11)$$

where, on the right hand side, the exponent $(\ldots)^t$ denotes elevation of a symmetric matrix to the power $t$. Note that

$$\mathbf{P}\#_0\mathbf{Q} = \mathbf{P} \qquad \mathbf{P}\#_1\mathbf{Q} = \mathbf{Q} \qquad (4.12)$$

The RA-FP algorithm is defined as follows. When $\Sigma_k$ is given, instead of of defining $\Sigma_{k+1}$ by (4.10), let

$$\Sigma_{k+1} = \Sigma_k \#_{t_k} f(\Sigma_k), \qquad (4.13)$$

where $t_k \in [0, 1]$. The RA-FP algorithm (4.13) is indeed a generalization of the FP algorithm (4.10), since putting $t_k = 1$ in (4.13) yields (4.10), as can be seen from (4.12). In our work, we set

$$t_k = \frac{1}{k+1} \qquad (4.14)$$

Similarly to ML-FS a pseudo-code description of the RA-FP algorithm is given in Algorithm 3 below.

---

**Algorithm 3** RA-FP

1: **Input**: $\mathbf{X} \in \mathbb{R}^{K \times T}$, optionally $\beta$
2: Initialize $\Sigma$ using either MoM or $\Sigma = \mathbf{I}_p$
3: If $\beta$ is not given initialize both $\Sigma$ and $\beta$ using MoM
4: **while** $(D(k) > \text{tol})$ and $(k < N_{\max})$ **do**
5:     Estimate $\Sigma$ using one iteration of (4.13)
6:     **if** $\beta$ is not given **then**
7:         Estimate $\beta$ by applying Newton-Raphson into (4.4)
8:     **else**
9:         Go to step 5
10:     **end**
11: **end**
12: Using $\Sigma$ and $\beta$, estimate $m$ with (4.3)
13: **Output**: $\Sigma, \beta, m$

---

In Appendix (A) we present a mathematical proof of the convergence of the RA-FP algorithm.

### 4.1.3 Experimental results

To quantify the performance of ML-FS and RA-FP, we generate data according to [76], [36] with $\Sigma$ defined by

$$\Sigma(i,j) = \sigma^{|i-j|}, \ i,j = \{1,2,...p-1\}, \tag{4.15}$$

where $\sigma$ belongs to the interval $[0,1)$ and controls the correlations between the entries of the data. All results are averaged over 500 runs. For these types of experiments we used $p = 3$, $N = 10000$, and $\sigma$ has been uniformly selected from the range $(0.4, 0.6)$.

Fig. 4.1 shows the difference between the estimated and the original matrix $\Sigma$ as a function of shape parameter and for known value of $\beta$. The difference is measured by the Frobenius norm. It can be observed that for $\beta < 1$, RA-FP and ML-FP provide better results than the MoM and ML-FS, while for $\beta \geq 2$ RA-FP performs the best among these four algorithms. It is worth mentioning, that when we refer to the original matrix $\Sigma$, we do not imply the true fixed point. A fixed point is a ML estimate and in practice is not know. Thus, by original matrix $\Sigma$ we refer to a matrix that is in the neighborhood of the ML estimate and as the number of samples approaches infinity this matrix gets closer to the ML estimate. Fig. 4.2 displays the Frobenius norm between the estimated and the original scatter matrix, when $\Sigma$ and $\beta$ have been jointly estimated. When $\beta > 4$ RA-FP performs the best and for $\beta < 1$ the ML techniques perform better than MoM.

Fig.4.3 shows the comparison between the variances of the $\beta$ estimators generated from the MoM, ML-FP, ML-FS, and RA-FP estimators as well as the Cramer-Rao lower bound (CRLB), as a function of different values of $\beta$. CRLB can be obtained by inverting the Fisher information matrix derived by [93]. Numerical experiments [76], show the unbi-asedness of the estimator $\beta$, so by Fig.4.3, we observe that the performance of the ML-FS and RA-FP (overlap) is very close to the CRLB, illustrating the MLE efficiency. On the other hand, the other two methods have issues when $\beta$ moves away from one.

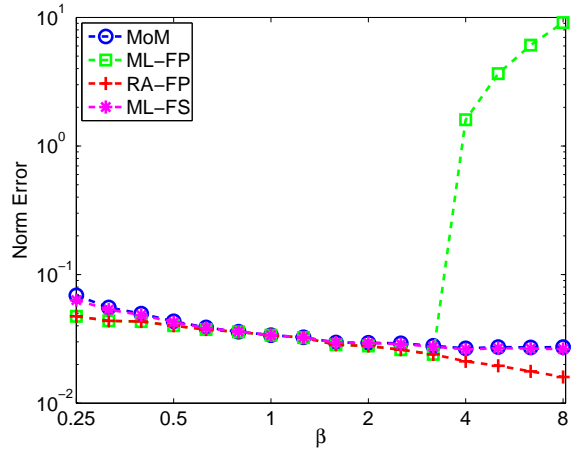

Fig. 4.1. Scatter matrix estimation performance for different values of shape parameter, for $N = 10000$, $\sigma \in (0.4, 0.6)$.

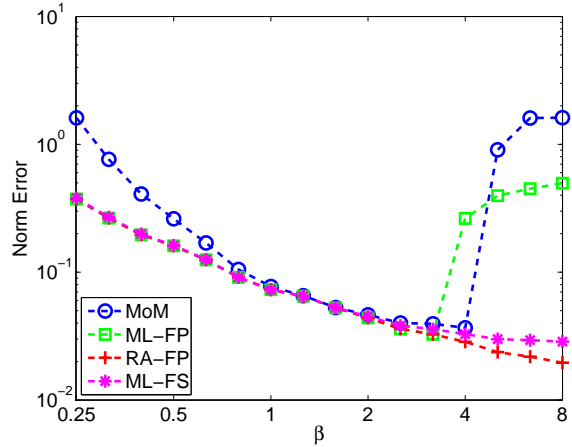

Fig. 4.2. Scatter matrix estimation performance for different values of shape parameter when $\Sigma$ and $\beta$ have been jointly estimated. $N = 10000$, $\sigma \in (0.4, 0.6)$.

Fig. 4.4 shows the difference between the estimated and the true $\Sigma$ as a function of $\beta$ for various initial values. As observed for any value of $\beta \in (0.25, 8)$ algorithm remains invariant to the initial point. This can result in the reduction of the computational cost of

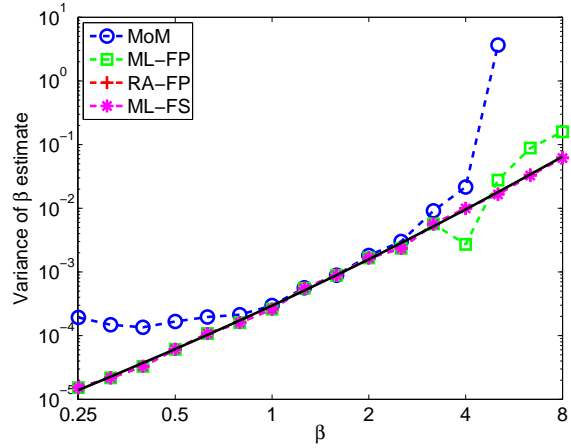

FIG. 4.3. Variance of $\beta$ estimate for different values of $\beta$. $T = 10000$, $\sigma \in (0.4, 0.6)$.

the algorithm, since addition computations produced by the MoM can be avoided. Finally,

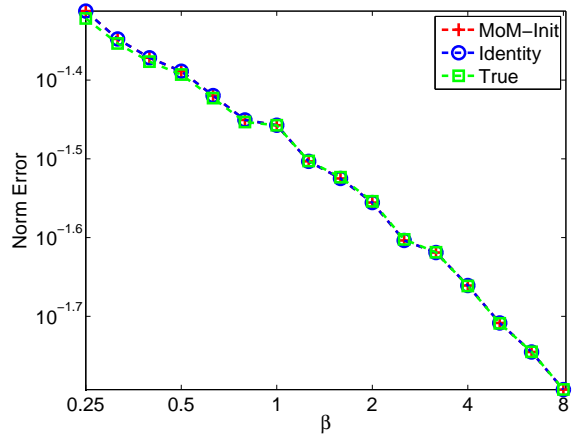

FIG. 4.4. Scatter matrix estimation performance for different values of shape parameter using different initial points of $\Sigma$. $N = 10000$, $\sigma \in (0.4, 0.6)$.

Fig. 4.5 shows the number of iterations for RA-FP to successfully converge. When $\beta = 1$ the MGGD reduces to the Gaussian distribution, where the ML estimator is the covariance matrix. Hence, only one iteration of RA-FP is needed. In addition, it can observed that the number of iterations depend on the value of $\beta$ and decreases as $\beta$ becomes large.

An interesting point that can be raised is, whether a usual $\epsilon$−perturbation applied to (4.10) can improve the local convergence over the ML-FP algorithm when $\beta \geq 2$. To illustrate this point, we implement $\epsilon$−perturbation as

$$\Sigma_{k+1} = (1 - \epsilon) f(\Sigma_k) + \epsilon \mathbf{I}, \tag{4.16}$$

where **Id** is the identity matrix, and $\varepsilon > 0$ is an arbitrary small constant. This iteration was applied to data generated using the parameters $p = 3$, $N = 1000$, and $\sigma = 0.5$. Fig. 4.6 shows the Frobenius distance between the estimated and true scatter matrices as a function of $\beta$. These results correspond to $\varepsilon = 0.01$, but similar results are obtained for $\varepsilon = 0.001$ and $\varepsilon = 0.0001$. It is clear that, for $\beta \geq 2$, ML-FP-eps (the above epsilon perturbation) does not provide local convergence, and behaves in the same way as ML-FP, (the standard fixed point algorithm).

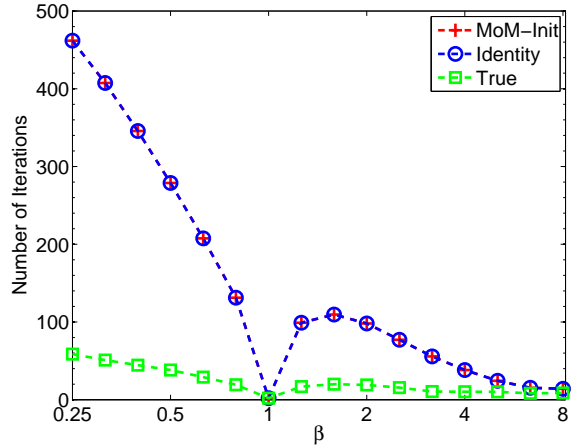

FIG. 4.5. Number of iterations needed for RA-FP to converge as a function of $\beta$. $N = 10000$, $\sigma \in (0.4, 0.6)$.

## 4.2  IVA-A-GGD

The high performance of the ML-FS and RA-FP algorithm for the estimation of $\Sigma$ and $\beta$ of an MGGD, motivates their use as the main estimation techniques for IVA. Similar to the ICA decoupled cost function given in (2.21), the IVA decoupled MI cost function depends on the differential entropy of the $n$th SCV. Therefore, by using the assumption that $\mathbf{y}_n$ is distributed according to an MGGD, the differential entropy of $\mathbf{y}_n$ becomes

$$H(\mathbf{y}_n) = -E\{\log p(\mathbf{y}_n)\} = -\log\left(\frac{\beta\Gamma\left(\frac{K}{2}\right)}{\pi^{\frac{K}{2}}\Gamma\left(\frac{K}{2\beta}\right)2^{\frac{K}{2\beta}}}\right) + \frac{K}{2}\log m + \frac{1}{2}\log|\Sigma| + \frac{1}{2m^\beta}E\left\{\left(\mathbf{y}_n^\top\Sigma^{-1}\mathbf{y}_n\right)^\beta\right\}.$$

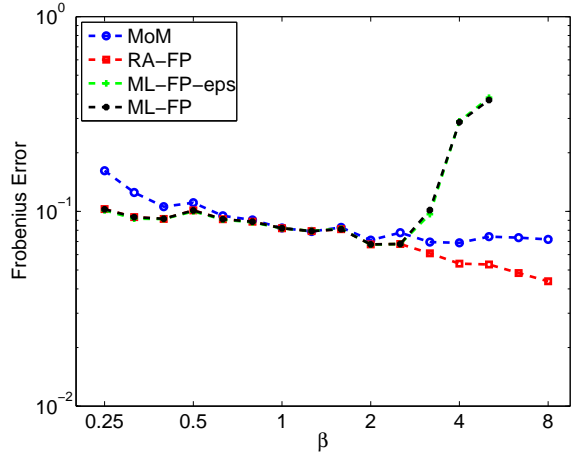

FIG. 4.6. Scatter matrix estimation performance for different values of the shape parameter

This allows us to rewrite (2.13) as a sequence of cost functions given by

$$J_{IVA} = -\log\left(\frac{\beta\Gamma\left(\frac{K}{2}\right)}{\pi^{\frac{K}{2}}\Gamma\left(\frac{K}{2\beta}\right)2^{\frac{K}{2\beta}}}\right) + \frac{K}{2}\log m + \frac{1}{2}\log|\Sigma| + \frac{1}{2m^\beta}E\left\{\left(\mathbf{y}_n^\top\Sigma^{-1}\mathbf{y}_n\right)^\beta\right\} - \log\left|\left(\mathbf{h}_n^{[k]}\right)^\top \mathbf{w}_n^{[k]}\right| - C_n^{[k]},$$

where $\mathbf{h}_n^{[k]}$ is the unit length vector, with the property that is perpendicular to all row vectors of the matrix $\mathbf{W}^{[k]}$ except of the vector $\mathbf{w}_n^{[k]}$ and is computed in a similar manner as it was for the non-orthogonal ICA framework. The term $C_n^{[k]}$ is a constant independent of $\mathbf{w}_n^{[k]}$. Therefore, the nonorthogonal IVA gradient is given by

$$\frac{\partial J_{IVA}}{\partial \mathbf{w}_n^{[k]}} = E\left\{\left(\frac{\beta}{m^\beta}\left(\mathbf{y}_n^\top\Sigma^{-1}\mathbf{y}_n\right)^{\beta-1}\Sigma^{-1}\mathbf{y}_n\right)\mathbf{x}^{[k]}\right\} - \frac{\mathbf{h}_n^{[k]}}{\left(\mathbf{h}_n^{[k]}\right)^\top . \mathbf{w}_n^{[k]}},$$

The pseudo-code description of the IVA with adaptive MGGD is presented in Algo-rithm 4.

---
**Algorithm 4** IVA with adaptive MGGD
---
1: **Input**: $\mathbf{X} \in \mathbb{R}^{N \times T \times K}$
2: For each $k = 1, ..., K$, initialize $\mathbf{W} \in \mathbb{R}^{N \times N}$
3: **for** $n = 1:N$ **do**
4:     Estimate scatter $\mathbf{\Sigma}_n$ and $\beta$ using either MoM, ML-FS or RA-FP
5:     Compute $\mathbf{h}_n^{[k]}$ for $k = 1, ..., K$, which are orthogonal to $\mathbf{w}_i^{[k]}$ for all $i \neq n$
6:     **for** $k = 1:K$ **do**
7:         Calculate the derivative $\frac{\partial J_{IVA}}{\partial \mathbf{w}_n^{[k]}}$
8:         $(\mathbf{w}_n^{[k]})^{\text{new}} \leftarrow (\mathbf{w}_n^{[k]})^{\text{old}} - \mu \frac{\partial J_{IVA}}{\partial \mathbf{w}_n^{[k]}}$
9:     **end** % $k$
10: **end** % $n$
11: $J_{IVA} = \sum_{n=1}^{N} H(\mathbf{y}_n) - \sum_{k=1}^{K} \log(|\mathbf{W}^{[k]}|)$
12: Repeat steps 3 to 11 until convergence in $\mathbf{W}$ or maximum iterations exceeded
13: **Output**: $\mathbf{W}$
---

The main part of this algorithm is the loop described in lines 3–11. The algorithm exits this loop when the relative difference between two successive estimates of each demixing matrix exceeds a pre-defined tolerance bound or whenever the number of iterations exceeds a pre-defined upper bound. Since the performance of an IVA algorithm highly depends on its ability to accurately estimate the distribution parameters of the latent sources, we generate three versions of the algorithm presented in (4). The only difference between these three algorithms is the manner in which the parameters of the MGGD are estimated. We call these three different algorithms IVA-A-GGD-MoM, IVA-A-GGD-RAFP, and IVA-A-GGD-MLFS.

## 4.3 Experimental results

To show the effectiveness of the IVA with adaptive MGGD, we compare its performance in terms of the joint inter-symbol-interference (ISI), defined in [6, 62]. We com-pare the different variants of our IVA with adaptive MGGD algorithms with six commonly used JBSS algorithms: IVA-GL is the IVA-L algorithm initialized by IVA-G. IVA-GGD is a more general IVA implementation that uses MoM to estimate the scatter matrix using a pre-defined list of shape parameters, such that it selects the one that provides the minimum cost function. Joint diagonalization using SOS (JDIAG-SOS) [59], uses symmetric orthogonal joint diagonalization of covariance matrices based on multiple datasets and ignores higher-order statistics. Joint diagonalization using HOS (JDIAG-HOS) [58] uses symmetric orthogonal joint diagonalization of fourth-order cumulants based and ignores second-order statistics. In this set of experiments, we generate MGGD sources and consider two different cases for the shape parameter $\beta$. For the first case we generate ten MGGD sources $N = 10$, $K = 3$, and $\beta$, $\sigma$ have been uniformly selected from the range $(0.25, 4)$ and $(0.4, 0.6)$ respectively. For the second case, we have $\beta \in (4, 8)$.

The performances are compared in terms of the inter symbol-interference (ISI). ISI [26, 71] is a widely used performance metric, which does not require an ordering of the sources in order to assess performance and is given by

$$\text{ISI}(\mathbf{G}) = \frac{\sum_{n=1}^{N}\left(\sum_{m=1}^{N}\frac{|g_{n,m}|}{\max_p |g_{n,p}|} - 1\right) + \sum_{m=1}^{N}\left(\sum_{n=1}^{N}\frac{|g_{n,m}|}{\max_p |g_{p,m}|} - 1\right)}{2N(N-1)},$$

where $\mathbf{G} = \mathbf{WA}$ is the global demixing-mixing matrix. In this work, we use average and joint ISI [7, 61] to measure the performance of the IVA algorithms. The average ISI is given by

$$\text{ISI}_{\text{AVG}}(\mathbf{G}^{[1]}, \ldots, \mathbf{G}^{[K]}) = \frac{1}{K}\sum_{k=1}^{K}\text{ISI}(\mathbf{G}^{[k]}),$$

where $\mathbf{G}^{[k]}$ for $k = 1, \ldots, K$ is the global demixing-mixing matrix for the $k$th data set. The

joint ISI is given by

$$\text{ISI}_{\text{JNT}}(\mathbf{G}^{[1]},\ldots,\mathbf{G}^{[K]}) = \text{ISI}\left(\frac{1}{K}\sum_{k=1}^{K}(\mathbf{G}^{[k]})\right) = \text{ISI}\left(\sum_{k=1}^{K}(\mathbf{G}^{[k]})\right).$$

We note that joint ISI takes the source alignment errors into account while average ISI does not.

From Fig.4.7, we observe that when the number of sample size is small, (*i.e.*, less than 1000), IVA-A-GGD-MoM, IVA-A-GGD-MLFS, and IVA-A-GGD-RAFP provide similar performance, while when the number of sample size increases IVA-A-GGD-RAFP performs the best among the seven algorithms since the large sample size allows for accurate estimation of of the shape parameter and scatter matrix. IVA-GGD does not show as accurate performance as the adaptive versions of IVA, due to the lack to precisely estimate the shape parameter of the underlying assumed MGGD. On the other hand JDIAG-SOS, JDIAG-HOS, and IVA-GL do not provide desirable performance due to the model mismatch. From Fig.4.8, we see that IVA-A-GGD-MLFS and IVA-A-GGD-RAFP provide similar performance, with IVA-A-GGD-RAFP being slightly better as the sample size increases. For the rest of the algorithms, we observe a similar trend with the results presented in Fig.4.7.

## 4.4 Summary

We present two new ML algorithms, ML-FS and RA-FP, which accurately estimate $\Sigma$ for any positive value of $\beta$. ML-FS is a variation of Newton-Raphson algorithm where the inverse Hessian has been replased by the inverse of the Fisher information matrix. RA-FP is based on the implementation of successful Riemannian averages of the FP iterates, in order to prevent them from diverging from the true value. Numerical results show that for any value of shape parameter, the sequence of matrices produced by ML-FS or RA-FP converges to the true value. Based on ML-FP and RA-FP, we introduce a class of IVA algorithms based on the assumption that each SCV is distributed according to an MGGD. The new algorithms, estimate the shape parameter and scatter matrix jointly, while taking both SOS and HOS into account. Experimental results, reveal the desirable performance of the new IVA algorithms when compared with existing competing algorithms.

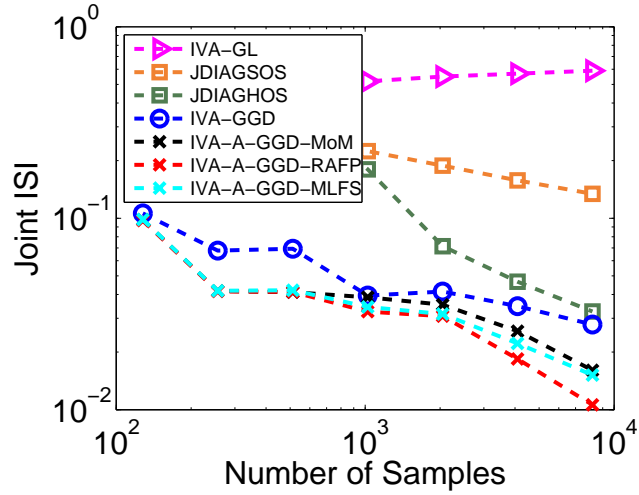

FIG. 4.7. Performance of IVA-A-GGD for different number of sample size and $\beta \in (0.25, 4)$.

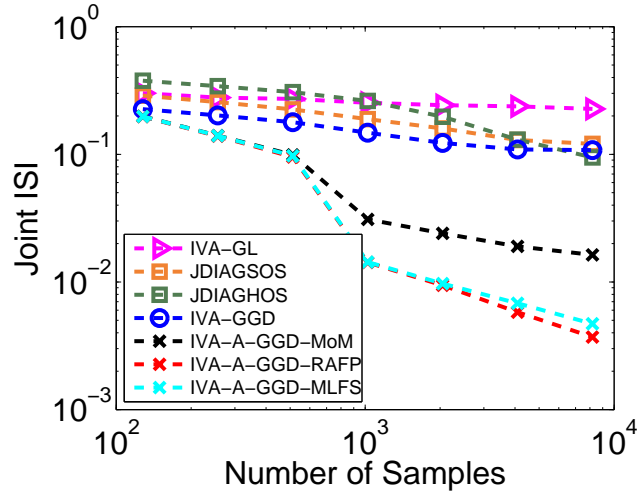

FIG. 4.8. Performance of IVA-A-GGD for different number of sample size and $\beta \in (4, 8)$.

# Chapter 5

# INDEPENDENCE AND SPARSITY:BALANCING TWO OBJECTIVES IN OPTIMIZATION FOR SOURCE SEPARATION

For a given dataset, BSS provides useful decompositions under minimum assumptions typically by making use of statistical properties—forms of diversity—of the data. Two popular forms of diversity that have proven useful for many applications are *statistical independence* and *sparsity*. Although many methods have been proposed for the solution of the BSS problem that take either the statistical independence or the sparsity of the data into account, there is no unified method that can take into account both forms of diversity simultaneously. In this chapter, we discuss a mathematical framework that enables direct control over the influence of these two forms of diversity and apply the proposed framework to the development of an effective ICA algorithm that can jointly exploit independence and sparsity. In addition, due to its importance in biomedical applications, we present a new model reproducibility framework for the evaluation of the proposed algorithm. Using simulated fMRI data, we illustrate the trade-offs between the use of sparsity versus independence in terms of the separation accuracy and reproducibility of the algorithm and provide guidance on how to balance these two objectives in real world applications where the ground truth is not available.

## 5.1 Sparsity- vs Independence-Based Methods

The objective of BSS methods is to decompose a set observations into the product of a mixing matrix and a matrix of latent sources. However, without the exploitation of any form of diversity, the matrix factorization problem is ill-posed. Two of the most popular forms of diversity that have proven useful in many practical applications and enable unique solutions up to scaling and permutation ambiguities are independence [1, 9, 12, 30, 44, 55, 70, 79] and sparsity [37, 67, 91, 91, 101].

It has been shown in previous chapters that ICA is a powerful method that solely relies on the independence of the sources and provides a unique decomposition such that the sources are statistically independent. In contrast, methods such as dictionary learning (DL) [91] and sparse component analysis (SCA) [18, 38], take the sparsity of the sources directly into account, yielding decompositions where the estimated components are as sparse as possible, subject to the same permutation and scaling ambiguities as ICA, however with uniqueness guarantees only under specific conditions [91].

By assuming that the observations can be expressed as sparse combinations of a dictionary $\boldsymbol{\Phi}$, DL seeks to estimate both the dictionary and the collection of weight vectors, $\mathbf{S}$, generally through an alternating estimation procedure. The cost for this task is given by

$$\min_{\boldsymbol{\Phi},\mathbf{S}} \|\mathbf{X} - \boldsymbol{\Phi}\mathbf{S}\|_F^2 + \lambda\|\mathbf{S}\|_{1,1}, \tag{5.1}$$

where $\|\mathbf{S}\|_{1,1} = \sum_{i=1}^{M} \sum_{j=1}^{N} |s_{ij}|$ and $\lambda$ is the regularization parameter. Different DL algorithms include those based on probabilistic learning methods, learning methods based on clustering, among others [91]. For a more detailed review of DL and its applications, we refer the reader to [91, 100].

A related method to DL that exploits solely sparsity is SCA. If $\boldsymbol{\Phi} \in \mathbb{R}^{K \times V}$ denotes a dictionary matrix, whose rows are called the atoms, then at the first step of SCA, $\boldsymbol{\Phi}$ is applied to the mixture matrix $\mathbf{X}$, to obtain $\mathbf{C_x} \in \mathbb{R}^{P \times K}$. In such a case, the column vectors $\mathbf{C_x}(k)$ $k = 1, \ldots K$, form the scatter plot $\{\mathbf{C_x}(k)\}_{k=1}^{K}$. If the dictionary has been selected properly, *i.e.*, has as sparse a representation of the data as possible, the elements of $\{\mathbf{C_x}(k)\}_{k=1}^{K}$ are almost aligned with the columns of the mixing matrix. In the second step, the mixing matrix $\mathbf{A}$ needs to be estimated by $\{\mathbf{C_x}(k)\}_{k=1}^{K}$. Thus, under the assumption that at most one source contributes to each point of the scatter plot, clustering techniques can be used to estimate $\hat{\mathbf{A}}$. The third step consists of the estimation of the source representations that can be denoted as $\mathbf{C_s} \in \mathbb{R}^{P \times K}$, due to the sparsifying transformation, $\mathbf{C_x} = \mathbf{X}\mathbf{\Phi}^\top$, that has been applied to the mixture matrix at the first step of SCA. Each column of $\mathbf{C_s}$ can be estimated through the minimization problem

$$\hat{\mathbf{C}}_\mathbf{s}(k) = \arg \min_{\mathbf{c}|\mathbf{C_x}(k)=\hat{\mathbf{A}}\mathbf{c}} \|\mathbf{c}\|_1, \tag{5.2}$$

where $\mathbf{c}$ is the vector that needs to be minimized such as $\mathbf{C_x}(k) = \hat{\mathbf{A}}\mathbf{c}$ and the solution of the minimization problem gives an estimate of the $k$th column of $\mathbf{C_s}$. The final step consists of reconstructing the sources by $\mathbf{y} = \mathbf{C_s}\mathbf{\Phi}$, when the initial dictionary matrix is orthogonal. For a more detailed discussion of SCA, we refer the reader to [38].

Although ICA, DL, and SCA have their own justifications in terms of the diversity they exploit, the differences among these methods do not facilitate transformation from one method to another, thus making it difficult to balance these two different forms of diversity. In order to take advantage of these two forms of diversity, jointly, many *ad hoc* methods have been proposed, such as by selecting a density model that favors sparse distributions in ICA as noted in [1] or by using sparsity transformations following ICA [65]. Although selecting the source distribution would allow the ICA model to enjoy the desirable large sample properties of the ML formulation [1, 30], the model would be limited to a specific type of sparse distribution [1]. Additionally, sparsity transformations are an indirect way of imposing sparsity and do not allow a direct way of controlling independence versus sparsity.

This motivates the development of a unified mathematical framework that can take into account both statistical independence and sparsity and enables direct control over the effect that independence and sparsity have on an optimization framework.

## 5.2 Sparse Independent Component Analysis

Classically, sparsity is measured using the $\ell^0$ norm, and is defined as the number of non-zero coefficients from a vector $\mathbf{u} \in \mathbb{R}^V$

$$\|\mathbf{u}\|_0 = \#\{u_i \neq 0; i = 1, \ldots V\}. \tag{5.3}$$

Although the incorporation of (5.3) into the ICA framework is the most direct way to impose sparsity on the ICA cost function, the $\ell^0$ norm is computationally intractable. On the other hand, the $\ell^1$ norm, defined as the sum of the absolute values of a vector's coefficients, has served as a computationally efficient sparsity regularizer see *e.g.*, [23, 82, 90]. For this reason, we propose a direct way to promote sparsity into the ICA model through the addition of an $\ell^1$ regularization term to the ICA cost function. The addition of this term is expected to improve separation performance beyond what is achieved solely through the maximization of independence when the underlying sources are truly sparse.

### 5.2.1 Cost Function

Balancing the contribution of sparsity for each of the individual sources while optimizing (2.4) is a difficult task, due to the $\log|\det(\mathbf{W})|$ term. This issue can be avoided by using the decoupling approach described in (2.4). Thus, expressing (2.4) as a sequence of

MI cost functions we have

$$J_{ICA}(\mathbf{w}_n) = H(y_n) - \log\left|\mathbf{h}_n^\top \mathbf{w}_n\right| - C_n, \quad n = 1, \ldots, N, \tag{5.4}$$

where $\mathbf{h}_n$ is a unit vector that is perpendicular to all row vectors of $\mathbf{W}$ except $\mathbf{w}_n$ and $C_n$ is a constant that contain all the terms that are independent of $\mathbf{w}_n$. Using (5.4), the proposed sequence of cost functions that take both independence and sparsity of each individual source into account is given by

$$J(\mathbf{w}_n) = J_{ICA}(\mathbf{w}_n) + \lambda_n f(y_n), \quad n = 1, \ldots, N, \tag{5.5}$$

where $f(y_n) = \|y_n\|_1$ is the regularization term and $\lambda_n$ is the sparsity parameter for $n = 1, \ldots N$. Note, that with a slight abuse of notation in (5.5), we treat $y_n$ as a vector where each coordinate corresponds to a sample drawn from the random variable $y_n$.. The $\ell^1$ norm is a non-differentiable function, so it is replaced by the the sum of multi-quadratic functions [56], given by

$$f(y_n) = \lim_{\epsilon_n \to 0} \sum_{v=1}^{V} \sqrt{y_{n_v}^2 + \epsilon_n}, \tag{5.6}$$

where $\epsilon_n$ is the smoothing parameter.

### 5.2.2 Algorithmic Development

ICA by entropy bound minimization (ICA-EBM) is a flexible and parameter-free algorithm that can maximize independence in an efficient manner [57]. It is due to this flexibility and ability to effectively maximize independence that ICA-EBM serves as the algorithm for the direct integration of (5.5).

The gradient of (5.5) with respect to (w.r.t) $\mathbf{w}_n$ is given by

$$\frac{\partial}{\partial \mathbf{w}_n} J(\mathbf{w}_n) = \frac{\partial J_{ICA}(\mathbf{w}_n)}{\partial \mathbf{w}_n} + \lambda_n \lim_{\epsilon_n \to 0} \sum_{v=1}^{V} \frac{y_{n_v}}{\sqrt{y_{n_v}^2 + \epsilon_n}} \mathbf{x}, \tag{5.7}$$

where

$$\frac{\partial J_{ICA}(\mathbf{w}_n)}{\partial \mathbf{w}_n} = -E\left\{\frac{\partial \log p(y_n)}{\partial y_n}\mathbf{x}\right\} - \frac{\mathbf{h}_n}{\mathbf{h}_n^\top \mathbf{w}_n},$$

and $p(y_n)$, can be adaptively determined for each estimated source independently. We refer to this new ICA algorithm as SparseICA-EBM. For better convergence properties, we follow the technique in [57] and define the domain of our cost function to be the unit sphere in $\mathbb{R}^N$. By using the projection transformation onto the tangent hyperplane of the unit sphere at the point $\mathbf{w}_n$, the normalized gradient of our cost function is given by

$$\mathbf{u}_n = \mathbf{P}_n(\mathbf{w}_n)\frac{\partial J(\mathbf{w}_n)}{\partial \mathbf{w}_n}, \tag{5.8}$$

where $\mathbf{P}_n(\mathbf{w}_n) = \mathbf{I} - \mathbf{w}_n\mathbf{w}_n^T$ and $\|\mathbf{w}_n\| = 1$.

In order to achieve fast convergence, SparseICA-EBM has been implemented using three stages. First, FastICA [42] is performed on the mixtures, generating an initial estimate of the demixing matrix $\mathbf{W}$. This estimate is further refined through the performance of orthogonal ICA using (5.5). The final stage consists of the application of non-orthogonal ICA using the estimated $\mathbf{W}$ obtained from the previous stage. The pseudo-code description of the non-orthogonal ICA stage is presented in Algorithm 5.

---

**Algorithm 5** SparseICA-EBM

1: **Input**: $\mathbf{X} \in \mathbb{R}^{N \times V}, \mathbf{W}_{\text{init}}, \lambda_n, \epsilon_n$
2: **for** $n = 1:N$ **do**
3:     Compute $\mathbf{h}_n$, orthogonal to $\mathbf{w}_i$ for all $i \neq n$
4:     Calculate the derivative $\frac{\partial J(\mathbf{w}_n)}{\partial \mathbf{w}_n}$ using (5.7)
5:     Project the gradient onto the unit sphere using (5.8)
6:     $(\mathbf{w}_n)^{\text{new}} \leftarrow (\mathbf{w}_n)^{\text{old}} - \gamma \mathbf{u}_n$
7: **end**
8: Repeat steps 2 through 7 until convergence in $J(\mathbf{W})$ or until the maximum number of iterations is exceeded
9: **Output**: $\mathbf{W}$

---

The term $J(\mathbf{W})$ introduced in Algorithm 5, is given by

$$J(\mathbf{W}) = \sum_{n=1}^{N} H(y_n) - \log|\det(\mathbf{W})| + \sum_{n=1}^{N} \lambda_n \|y_n\|_1. \qquad (5.9)$$

The new SparseICA-EBM not only provides flexible density matching but also yields solutions with variable levels of sparsity, through manual selection of $\lambda_n$ and $\epsilon_n$.

## 5.3 Simulated Sparse Data

We demonstrate the performance of SparseICA-EBM (5.5), in terms of its estimation accuracy, using simulated sparse sources. We compare the SparseICA-EBM algorithm with the original ICA-EBM algorithm and due to its popularity in many applications, we also compare SparseICA-EBM with two implementations of the Infomax algorithm [11]. One version is based on the natural gradient optimization framework (Infomax-NG) and the other one is based on a quasi-Newton technique Broyden, Fletcher, Goldfarb, and Shanno (BFGS) [73], which we call Infomax-BFGS.

### 5.3.1 Experiments and Numerical Results

We generate 20 simulated sources, each of which is distributed according to a GGD with sample size $T = 10^3$. The PDF of each source is given by [72]

$$p(x; \beta, \sigma) = \eta \exp\left(-\frac{x^{2\beta}}{2\sigma^{2\beta}}\right), \; x \in \mathbb{R}$$

where $\eta = \frac{\beta}{2^{\frac{1}{2\beta}} \Gamma(\frac{1}{2\beta}) \sigma}$. The shape parameter, $\beta$, controls the peakedness and spread of the distribution as well as its sparsity. If $\beta < 1$, the distribution is more peaky than the Gaussian with heavier tails, and if $\beta > 1$, it is less peaky with lighter tails. Thus, as $\beta \to 0$ the distribution becomes more sparse.

To verify the sparse nature of the sources used for the first set of the experiments, we generate 20 sources with sample size $T = 10^4$ and shape parameter $\beta$ from the range [0.1, 0.5] with a step size of 0.05. For each specific source, we measure the sparsity level using the Gini Index, defined as [41]

$$S(\mathbf{u}) = 1 - 2 \sum_{v=1}^{V} \frac{u^{(v)}}{\|\mathbf{u}\|_1} \left( \frac{V - v + 1/2}{V} \right),  \tag{5.10}$$

where $u^{(1)} \leq u^{(2)} \leq \cdots \leq u^{(V)}$ are the ordered coordinates of the vector $\mathbf{u} \in \mathbb{R}^V$ and average over the sources that correspond to a specific $\beta$. Note form (5.1) that Gini index is normalized, with 1 corresponding to very sparse sources while 0 to dense sources. In Fig. 5.1, we see that as we increase $\beta$, sources become less sparse.

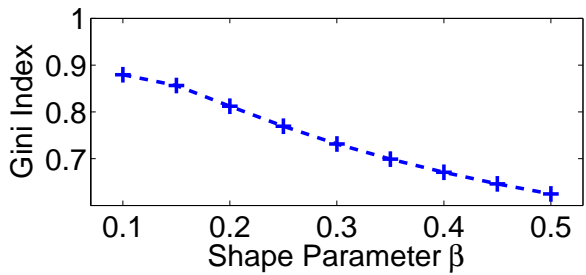

FIG. 5.1. Average Gini Index as a function of the shape parameter, $\beta$. The Gini Index is normalized and 1 corresponds to very sparse sources while 0 to dense sources.

To evaluate the performance of the algorithms, we use the ISR as in [60]. For SparseICA-EBM, the algorithm parameters are $\lambda = 10^4$ and $\epsilon = 10^{-2}$ and are determined based on a grid search selection. All results are the average of 300 independent runs.

In Fig. 5.2, we display the normalized ISR as a function of $\beta$. We observe that for small values of $\beta$, *i.e.*, highly sparse case, SparseICA-EBM exhibits better performance. On the other hand, ICA-EBM starts performing better than the other algorithms as we increase $\beta$, *i.e.*, decrease sparsity. It is worth mentioning that Infomax-NG often fails to converge as $\beta$ increases revealing its poor performance under this experimental setup. On the other hand, Infomax-BFGS shows reasonable performance especially for small values of $\beta$.

In Fig. 5.3, we display the normalized ISR as a function of the sample size. To study the case where sources are very sparse we generate all sources using $\beta = 0.1$. As the sample size increases, SparseICA-EBM and ICA-EBM perform better than the other two algorithms, since the large sample size enables an accurate approximation of the differential entropy of the estimated sources. When the sample size becomes greater than $10^3$, Infomax-BFGS starts providing highly inaccurate results, due to algorithmic issues in the approximation of the inverse of the Hessian matrix.

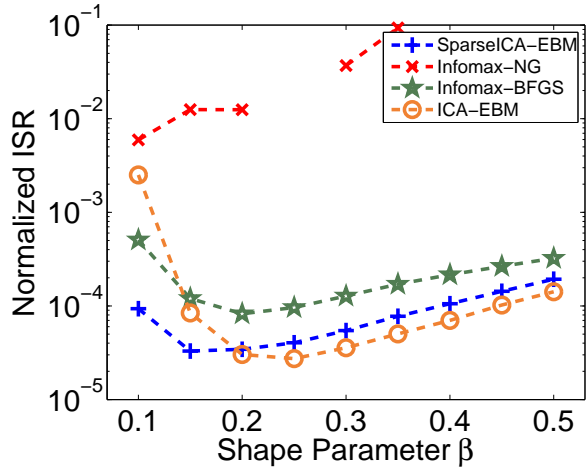

Fig. 5.2. Performance comparison of four ICA algorithms in terms of the normalized average ISR as a function of shape parameter, $\beta$, for 20 sources with $T = 10^3$. Each point is the result of 300 independent runs.

Finally, in Fig. 5.4, we display the normalized ISR as a function of the number of sources where for each source $T = 10^3$ and $\beta = 0.1$. It is clear from Fig. 5.4 that SparseICA-EBM shows the best performance. Infomax-BFGS performs well when the number of sources is small since the optimization procedure is performed in a low dimensional space. This reveals the benefit of employing the decoupling approach, since the reduction to a set of vector optimization problems avoids over-complicated surfaces for the cost function.

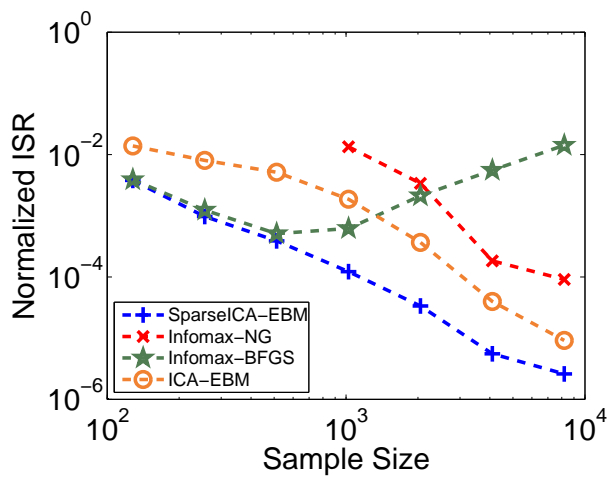

FIG. 5.3. Performance comparison of four ICA algorithms in terms of the normalized average ISR as a function of sample size, $T$, for 20 sources with $\beta = 0.1$. Each point is the result of 300 independent runs.

## 5.4 Simulated fMRI Data

For a BSS algorithm to be useful in real world applications such as the analysis of fMRI data, it must be able to efficiently extract the latent sources and do so consistently. Consequently, motivated by [87], to evaluate our proposed model, we consider two different metrics of performance. The first is in terms of its estimation accuracy, *i.e.*, its ability to accurately extract the latent sources, and the second is in terms of its reproducibility, *i.e.*, the consistency of the solutions across different datasets and runs. Such metrics are especially important for the analysis of fMRI data, since if sources are extracted incorrectly, the conclusion may be flawed, for instance leading to improper identification of biomarkers, *i.e.*, spatial patterns, of disease. Generally when using ICA on fMRI data, the estimated components tend to have sparse distributions [22], motivating the study of the synergy between independence and sparsity. Therefore, using these two measures of performance, we explore the tradeoffs between the use of sparsity versus independence, through fMRI data, and provide a guidance on how to balance these two objectives in real world applications

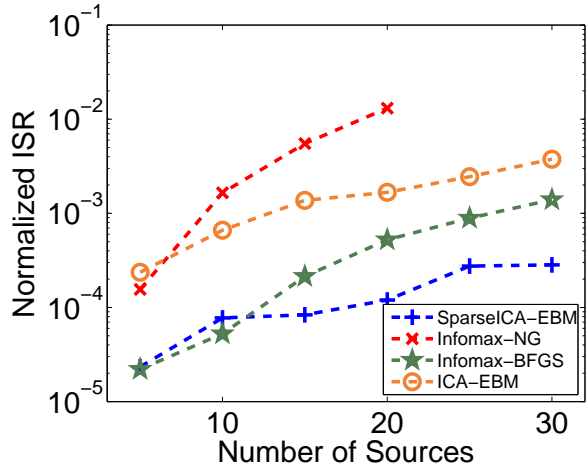

FIG. 5.4. Performance comparison of four ICA algorithms in terms of the normalized average ISR as a function of number of sources, $N$, with $T = 10^3$ and $\beta = 0.1$. Each point is the result of 300 independent runs.

where the ground truth is not available.

This investigation is performed through the generation of simulated fMRI data using SimTB [33], which enables flexible generation of fMRI–like datasets under a model of spatio-temporal separability. To study the effect of independence against that of sparsity, we generate 10 datasets, each representing a different subject with 20 sources, for three different scenarios each with different levels of noise. The three scenarios are shown in Fig. 5.5 and consist of the cases where all sources are very sparse with little to no spatial overlap, a mixture of very sparse and less sparse sources again with little to no spatial overlap, and very sparse as well as less sparse sources with an increased amount of spatial overlap. The sparsity and the degree of overlap of the original sources is controlled by adjusting the SimTB parameter value that controls the "spread" of the sources. Note that when we decrease the spread of each individual source the sparsity of this particular source is decreased. This comes from the definition of a sparse distribution which is one for which most of the energy is contained in only a few of the coefficients [41]. The additive noise is Rician distributed and has energy specified by the contrast-to-noise ratio (CNR) defined as the ratio of the temporal standard deviation of the true signal divided by the temporal standard deviation of the noise [33]. Each source is a $100 \times 100$ image and the length of the experiment is 260 samples, meaning that simulated **X** is of dimension $260 \times 10^4$.

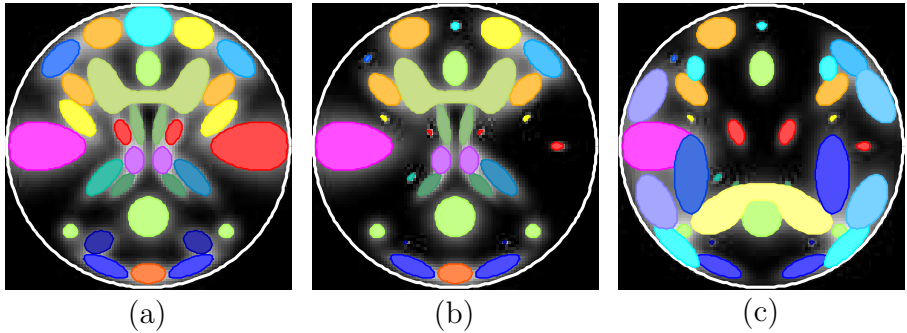

FIG. 5.5. Simulated fMRI-like components for the three different scenarios. Note that each color indicates a different component. The scenarios are (a) all sources are very sparse with no spatial overlap, (b) a mixture of very sparse and less sparse sources and no spatial overlap, (c) very sparse as well as less sparse sources with a certain degree of spatial overlap.

The average Gini indices for the 20 sources and for the three different scenarios are summarized in Fig. 5.6 (a). Additionally, we compute the average correlation across subjects and display the distribution of the values in Fig. 5.6 (b). Note that the mean and standard deviation of the pairwise source correlations are: $0.044 \pm 0.034, 0.022 \pm 0.031$, and $0.03 \pm 0.043$, respectively.

### 5.4.1 Balancing Independence and Sparsity

Since the ground truth is available for our simulated sources, we evaluate the performance of SparseICA-EBM in terms of its estimation accuracy, using the average absolute value of the correlation between the true and the estimated sources. Thus, for the first part of our study, we evaluate the correlation coefficient between the true and the estimated spatial maps as a function of $\lambda_n$ and $\epsilon_n$. Since $\lambda_n$ controls the degree to which sparsity is em-phasized over independence in SparseICA-EBM, we would like to visualize the behavior of the algorithm when we relax the independence assumption for each of the three groups and for different levels of noise.

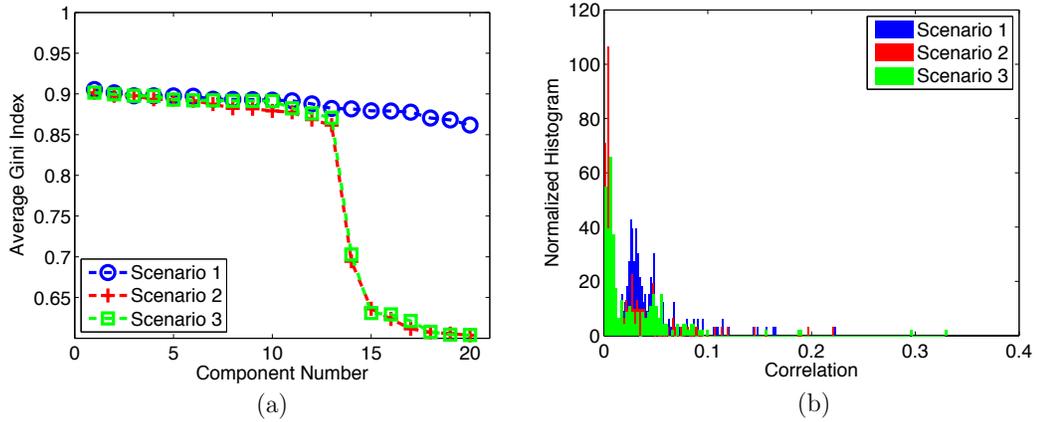

FIG. 5.6. (a) Average Gini index (b) distribution of the correlation values of the 20 latent sources for the three different scenarios. The Gini Index is normalized, with 1 corresponding to very sparse sources while 0 to dense sources.

The first step in processing the fMRI-like data consists of the application of PCA to each dataset, individually. Since 20 sources are generated for each dataset, the dimension of each dataset is temporally reduced to 20. After dimension reduction, we apply SparseICA-EBM to each dataset. After SparseICA-EBM has been applied to each subject's data, we pair the extracted components with the true latent sources. In the case where more than one estimated component is paired with a single true source, we use the Bertsekas algorithm [13], an iterative method that maximizes a given cost in a bipartite graph, to find the best assignment.

### 5.4.2 Model Reproducibility

Although, estimation accuracy is an effective metric to evaluate the separation power of a BSS algorithm, in many applications such as the analysis of fMRI data, model re-producibility is an important performance metric. Its importance derives from its ability to reveal how consistently an algorithm can produce similar estimated sources across different sets of data that are supposed to have come from the same distribution, such as different scans of the same subject. Thus, we also study the reproducibility of the SparseICA-EBM as a function of the sparsity parameter $\lambda_n$ and the smoothing parameter $\epsilon_n$. Motivated by the the nonparametric, prediction, activation, influence, reproducibility, resampling (NPAIRS) framework in neuroimaging [87], we split the original dataset into two, and perform separate analyses on each of the sub-datasets and study the similarity of the two sets of resulting separated sources. Since selecting certain rows of **X** is equivalent to sub-sampling the corresponding rows of **A** multiplied by the source matrix **S**, the similarity of the estimated sources is a good measure of the reproducibility of the proposed algorithm. A graphical illustration of this approach is presented in Fig. 5.7.

For this analysis, we split the mixture matrix **X**, defined as the collection of all realizations of **x**($v$), into two submatrices by selecting every other row of **X** creating $\mathbf{X}_1$ and $\mathbf{X}_2$, for each of the subjects. We apply PCA to each $\mathbf{X}_1$ and $\mathbf{X}_2$ for each subject and reduce their dimension to 20. After dimension reduction, we apply SparseICA-EBM to each reduced dataset. After SparseICA-EBM has been applied to the reduced submatrices, we pair the extracted components from the first submatrix with the extracted components from the second submatrix for each subject. In the case of multiple assignments, we again use the Bertsekas algorithm to determine the optimal assignment. We measure how close the pairs of estimated components are using the average absolute value of the correlation across subjects.

### 5.4.3 Experimental Results

Fig. 5.8 displays the average spatial correlation between the true and the estimated components as a function of the two key parameters for SparseICA-EBM, the regulariza-tion parameter $\lambda_n$ and the smoothing parameter $\epsilon_n$. Fig. 5.9 displays the average spatial correlation between the estimated components generated when applying SparseICA-EBM on the first half and the other half of the data as a function of $\lambda_n$ and $\epsilon_n$. For both figures the first column shows the results where data have been generated with no noise and the second column when noise has CNR = 1. For each noise level, we show the behavior and the reproducibility of the algorithm for the three different scenarios as described in the previous section. The hardware used in the computational studies is part of the UMBC High Performance Computing Facility (HPCF), for more information see hpcf.umbc.edu. Note that since the effectiveness of a BSS algorithm depends on both its accuracy and its consistency, in the two sets of figures that we present, we seek to find values of $\lambda_n$ and $\epsilon_n$ for which we obtain high source reconstruction accuracy as well as high reproducibility.

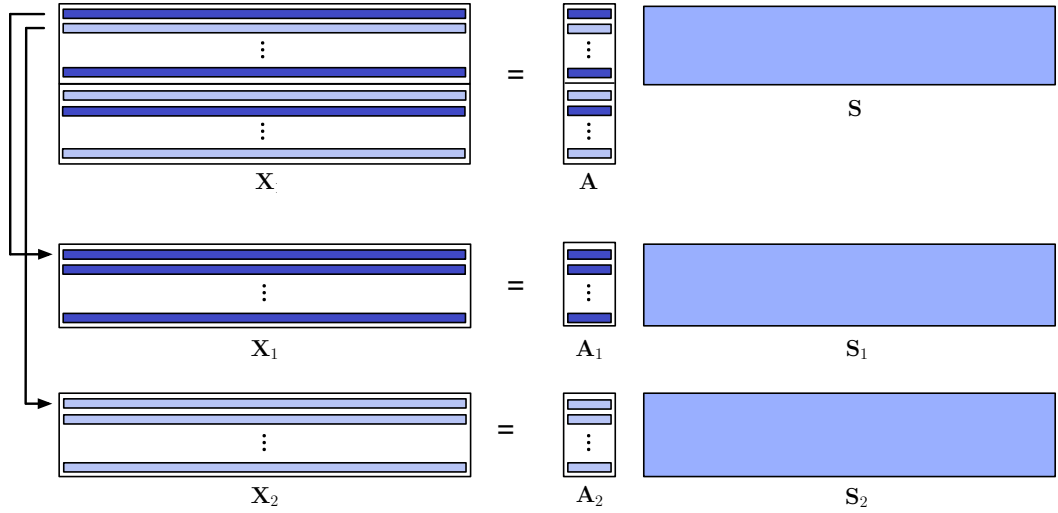

FIG. 5.7. Visualization of subsampling method used to split the observation matrix in order to evaluate the reproducibility of the model. Note that under this reproducibility framework $\mathbf{S}_1 \cong \mathbf{S}_2 \cong \mathbf{S}$.

From Fig. 5.8(a) and (d), we observe that when the original sources do not have significant overlaps and all of them are characterized as very sparse, a high value of $\lambda_n$ and $\epsilon_n$ produces higher average spatial correlation values, with the gain decreasing when noise level increases. This result shows that, in the case of truly sparse and independent sources, promoting sparsity within an ICA framework improves performance, since we effectively exploit another form of diversity, *i.e.*, property of the sources. In Fig. 5.8(b), we observe that, when some of the sources are sparse and some are less sparse, for high values of $\epsilon_n$, SparseICA-EBM with sparsity enforced, *i.e.*, high values of $\lambda_n$, provides better results, than with small values of $\lambda_n$, since only a third of the total sources are less sparse, thus the performance is dominated by the extraction of the sources that are sparse. From Fig. 5.8(e), SparseICA-EBM with only independence enforced, *i.e.*, small values of $\lambda_n$, and SparseICA-EBM with sparsity enforced and high values of $\epsilon_n$ provide similar separation performance, since the additive noise destroys the sparse nature of the data. Finally from Fig. 5.8(c) and (f), SparseICA-EBM with high values of $\lambda_n$ and $\epsilon_n$ provides similar results to SparseICA-EBM with low values of $\lambda_n$.

From Fig. 5.9(a) and (d), we observe that when the original sources do not overlap and all of them are characterized as very sparse, high values of $\lambda_n$ and for almost all values of $\epsilon_n$ the results are highly reproducible. Thus, for these cases, SparseICA-EBM produces both accurate and consistent results for large values of $\lambda_n$ and $\epsilon_n$. A similar trend can be observed in Fig. 5.9(b), where some of the sources are very sparse and some are less sparse. Fig. 5.9(e), for all values of $\lambda_n$ and $\epsilon_n$, SparseICA-EBM is becoming always consistent. Fig. 5.9(c), shows that, for some intermediate values of $\epsilon_n$, we have high reproducibility. Finally, in Fig. 5.9(f), SparseICA-EBM shows nearly identical results except for high values of $\lambda_n$ and $\epsilon_n$.

Based on Figs. 5.8 and 5.9, we can draw several interesting conclusions regarding the behavior of SparseICA-EBM, also can note few points for the selection of its parameters when we are working with real fMRI data. Since our goal is to have both high performance and high reproducibility, we observe that for the first and second scenarios where component overlaps are limited, sufficiently high values of $\lambda_n$, *i.e.,* in the interval $(10^{-2}, 10^4)$, as well as sufficiently high values of $\epsilon_n$, *i.e.,* in the interval $(0.5, 10)$, will produce sparse and smooth sources consistently. Moreover, for scenario 1, SparseICA-EBM with very small $\lambda_n$ is robust to noise. For the third scenario and when the values of $\lambda_n$ are small, SparseICA-EBM has relatively high performance. Therefore, for real world applications where all or a majority of sources can be assumed to be sparse high values of $\lambda_n$ and $\epsilon_n$ are expected to provide reasonable results, consistently. However for the case where overlaps are likely, by emphasizing both independence and sparsity in the optimization procedure will produce better overall performance.

An additional point worth noting, is that from Fig. 5.9(a) we observe a significant drop in reproducibility for $\lambda_n = 10^{-3}$. The SparseICA-EBM cost function consists of an independence term that is described by the negative of the ICA maximum likelihood function and a sparsity term that is described by the $\ell^1$ norm. Since the contribution of sparsity is weighted by the parameter $\lambda_n$ and the optimal solutions of the two terms are not necessarily the same, changing the value of $\lambda_n$ affects the overall solution space each time SpaceICA-EBM is applied to $\mathbf{X}_1$ and $\mathbf{X}_2$. To illustrate this point we perform the following experiment. To illustrate this point we measure the amount of sparsity and independence in the optimization procedure as a function of different values of $\lambda_n$ for a fixed $\epsilon_n = 1.46$. Every time we apply SparseICA-EBM to either $\mathbf{X}_1$ or $\mathbf{X}_2$, we calculate the magnitude of the distance between the sparsity term in the cost function and the overall cost function at the last iteration of SparseICA-EBM and denote it by $d_S$. Similarly, we calculate the magnitude of the distance between the independence term in the cost function and the overall cost function at the last iteration of SparseICA-EBM and denote it by $d_I$. Then we calculate the ratio of $d_S$ to $d_I$, (*i.e.* $r_1 = d_S/d_I$). We follow the same procedure when SparseICA-EBM is applied to $\mathbf{X}_2$ and calculate $r_2$. Thus, we generate two ratios $r_1$ and $r_2$ for each value of $\lambda_n$. From Table. 5.1, we observe that $r_1$ and $r_2$ are getting close to 1, as $\lambda_n$ approaches the value that corresponds to the drop on the surface shown in Fig. 5(a). This implies that the sparsity and independence terms contribute almost equally in the optimization procedure. This phenomenon expands the solution space, resulting in more local optima, and thus, when SpaceICA-EBM is applied to $\mathbf{X}_1$ and $\mathbf{X}_2$ separately, it yields pairs of estimated components that correlate less with each other. The performance drop observed in Fig. 5(a) would start to disappear in the rest of the figures for which noise is introduced to the data, since noise destroys the sparsity, or for scenarios where we manually reduce the sparsity of the original sources. Since sparsity is insufficient to fully extract the sources, the solution space of the second term in the cost function is close to being flat. This results in a joint cost surface with fewer local minima and therefore better correlation between the two sets of estimated components.

|       | $\lambda_n^1$ | $\lambda_n^2$ | $\lambda_n^3$ | $\lambda_n^4$ | $\lambda_n^5$ | $\boldsymbol{\lambda_n^6}$ | $\lambda_n^7$ | $\lambda_n^8$ | $\lambda_n^9$ | $\lambda_n^{10}$ | $\lambda_n^{11}$ | $\lambda_n^{12}$ | $\lambda_n^{13}$ |
|---|---|---|---|---|---|---|---|---|---|---|---|---|---|
| $r_1$ | $10^4$ | $10^3$ | 480.39 | 48.21 | 4.56 | **0.47** | 0.03 | 0.003 | $10^{-4}$ | $10^{-5}$ | $10^{-6}$ | $10^{-7}$ | $10^{-8}$ |
| $r_2$ | $10^4$ | $10^3$ | 480.54 | 48.23 | 4.56 | **0.47** | 0.03 | 0.003 | $10^{-4}$ | $10^{-5}$ | $10^{-6}$ | $10^{-7}$ | $10^{-8}$ |

Table 5.1. The two ratios $r_1$ and $r_2$ for each value of $\lambda_n$ and for a fixed $\epsilon_n$. For the value $\lambda_n^6$ we observe the performance slot in Fig. 5(a).

## 5.5 Conclusion

Methods that exploit sparsity and independence have proven useful in many applications. This motivates the development of a method that can effectively take into account both forms of diversity. In this chapter, we present a new mathematical framework that enables direct control over the influence that independence and sparsity have on the result and use this framework to generate a powerful algorithm that takes both sparsity and independence into account. We explore the tradeoffs between emphasizing these two objectives for different scenarios of simulated fMRI data and provide a guideline on the parameter selection for fMRI analysis when the ground truth is not available. Our results indicate that careful selection of the regularization parameters under certain scenarios will increase the quality of the final extracted sources enabling meaningful interpretations for fMRI analysis.

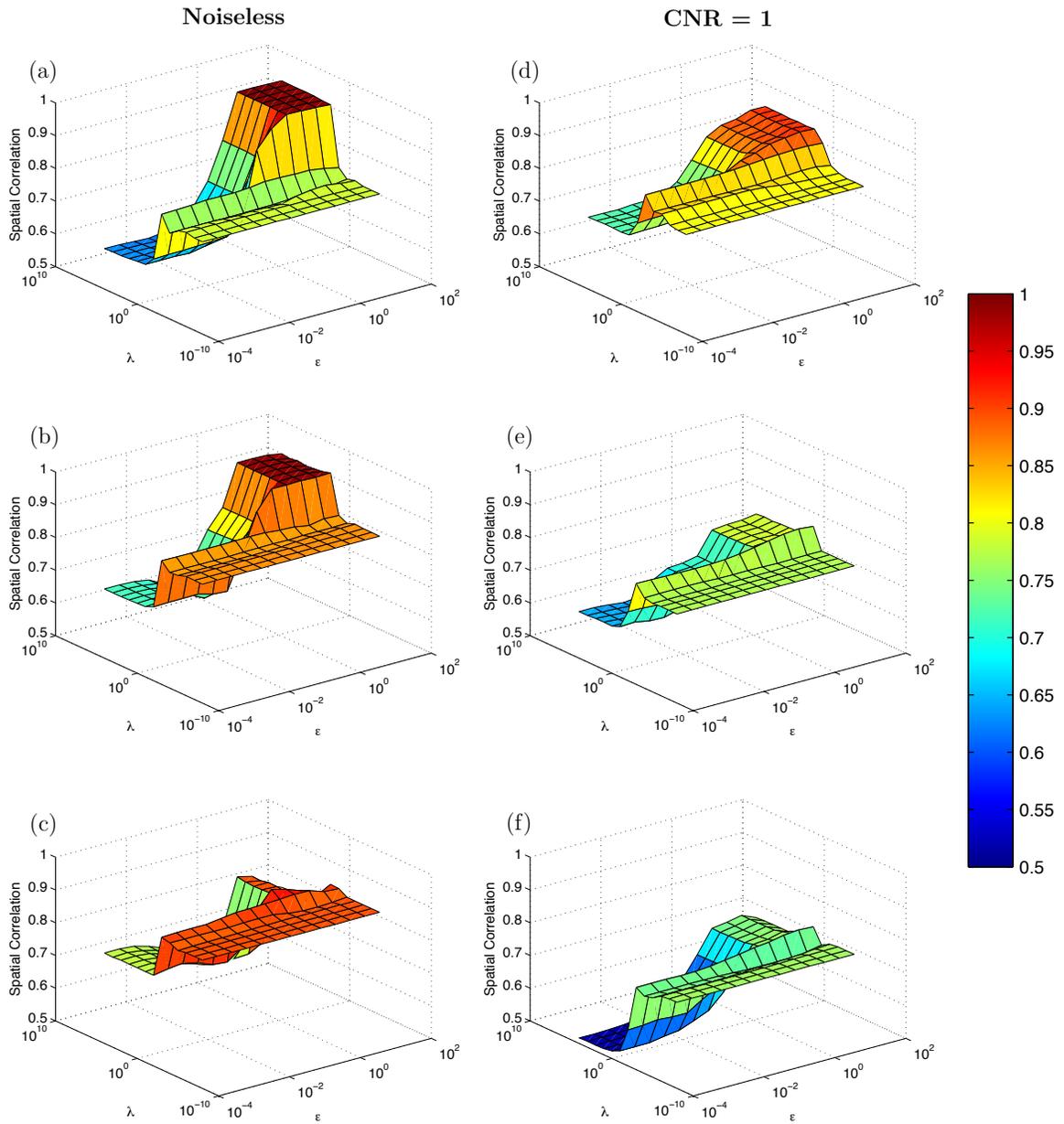

FIG. 5.8. Spatial correlation of the true and the estimated sources as a function of $\lambda_n$ and $\epsilon_n$ for different CNR values: (a)-(c) is the noiseless case and (d)-(f) have a CNR of 1. Plots (a) and (d) are from scenario 1, all sparse sources. Plots (b) and (e) are from scenario 2, some sparse and some less sparse sources with no overlap. Plots (c) and (f) are from scenario 3, some sparse sources and some less sparse sources with some degree of overlap. The results are the average of 128 runs.

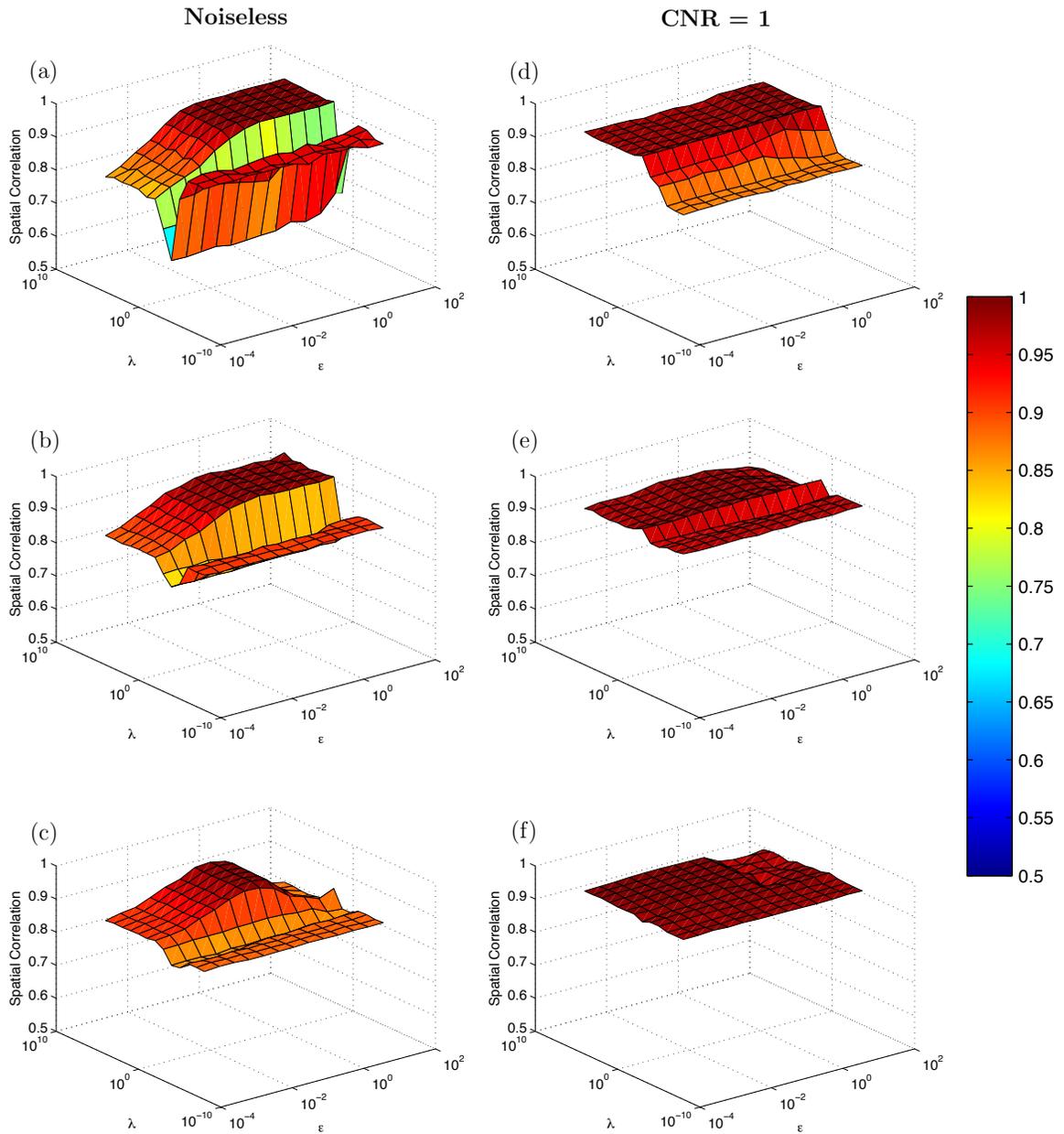

FIG. 5.9. Spatial correlation of the estimated components generated when applying SparseICA-EBM on the two halves of the data as a function of $\lambda_n$ and $\epsilon_n$ for different CNR values: (a)-(c) is the noiseless case and (d)-(f) have a CNR of 1. Plots (a) and (d) are from scenario 1, all sparse sources. Plots (b) and (e) are from scenario 2, some sparse and some less sparse sources with no overlap. Plots (c) and (f) are from scenario 3, some sparse sources and some less sparse sources with some degree of overlap. The results are the average of 64 runs.

# Chapter 6

# CONCLUSIONS AND FUTURE WORK

In this dissertation, we provide several important developments for modeling a signal and we consider applications of these models to the BSS problem. Our work enables increased flexibility in our developed BSS algorithms and, by utilizing additional forms of diversity, the increase in these algorithms' separation power over existing algorithms. In this chapter, we summarize our work and suggest several possible directions for further research.

## 6.1 Summary

As it has been mentioned before, formulating the ICA problem using an ML framework enables one to exploit many forms of diversity of the dataset described through its statistical properties. This requires knowledge of the true underlying PDF of the latent sources, which is unknown in many applications. In this dissertation, we present a new and efficient ICA algorithm, ICA-EMK, that utilizes both global as well as *adaptive* local measuring functions to gain insight into the local behavior of the source PDFs with only a modest increase in model complexity. By taking advantage of the decoupling trick, the optimization procedure of ICA-EMK is performed in a parallel fashion, thus allowing the computation time to become not only a function of the number of sources, but also proportional to the number of available processing cores. Experimental results confirm the attractiveness of the new ICA algorithm to separate sources from a wide range of distributions.

IVA is a generalization of ICA that makes full use of the statistical dependence across multiple datasets to achieve source separation, and can take both SOS and HOS into account. The MGGD provides an effective model for IVA and the performance of the IVA algorithm highly depends on the estimation of the scatter matrix and the shape parameter. The performance of existing approaches significantly suffers when the value of shape parameter becomes large, which makes them unsuitable for many applications. In this dissertation, we present ML-FS and RA-FP algorithms for the accurate estimation of the scatter matrix for any value of the shape parameter. We integrate both techniques into IVA to precisely estimate all the parameters of the MGGD sources simultaneously, which in turn leads to effective calculation of the IVA score and cost functions, resulting in better IVA performance.

Though statistical independence is a natural assumption in many cases, there are many practical applications where such a strong assumption is unrealistic. Often in these cases, some important prior information such as the sparsity about the data is available and incorporating it into the ICA model will result in better overall separation performance. In this dissertation, we propose a new mathematical framework that enables direct control over the influence that independence and sparsity have on the result and use this framework to generate a powerful ICA algorithm that takes both sparsity and independence into account. We explore the trade-offs between emphasizing these two objectives for different scenarios of simulated fMRI data and provide a guideline on the parameter selection for fMRI analysis when the ground truth is not available.

## 6.2 Future Directions

The main contribution of this dissertation is the development of effective and flexible ICA and IVA algorithms. The desirable performance of these proposed algorithms motivates several promising future theoretical and algorithmic developments as well as their use in several novel applications. We expand upon the directions for future research below.

### 6.2.1 Algorithmic Developments

Using our work as a starting point, we propose to extend the algorithmic capabilities of ICA and IVA algorithms in three major directions: the design of a more general multivariate PDF estimator, the investigation of independent vector subspace analysis, and the estimation of optimal values for the sparsity parameters of SparseICA-EBM.

**Maximum joint entropy densities**  Current IVA algorithms are based on the assumption that the underlying density model of the sources is unimodal and symmetric as well as that the samples are i.i.d. These assumptions are often not realistic and can lead to poor separation performance. Motivated by the flexibility and superior performance that ICA-EMK provides in the univariate case, we propose the design of a multivariate PDF estimator based on the maximum entropy principle to successfully match multivariate sources from a wide range of distributions. In addition, sample dependence can be taken into account through the use of an AR model in a similar manner as described in [35]. The proposed approach is quite different from the MGGD-based approach described in our work, where the MGGD model is chosen and the parameters are estimated during the estimation of the demixing matrix. The main challenges in the multivariate case that one can face are the following:

- Choice of multivariate measuring functions, to ensure that we provide flexible density

estimation while keeping the computational complexity low;

- Choice of the number and type of constraints;

- Multi-dimensional integration during the estimation of the Lagrange multipliers;

- Estimation of the AR coefficients.

**Independent vector subspace analysis**  As a motivation for the introduction of prior information into the IVA model, one can think of the example of analysis of data sets with a group structure, such as multiple fMRI, or fMRI and electroencephalography (EEG) data sets. Within these sets, it is to be expected that multiple highly correlated components may appear, all of these dependent on a set of components in each of the other data sets. This is the case, for example, when studying the expressions of tasks on a group of subjects, where natural subgroups can be formed based on age, gender, or other subject-specific parameters. Under these conditions, it is highly probable that subjects within a subgroup respond similarly to a task and thus to find multiple similar expressions that differentiate between the subgroups. Hence, the expressions could no longer be regarded as independent components but should be regrouped so as to form independent subspaces. The IVA formulation offers a number of unique advantages. It makes use of the powerful assumption of independence and it has a simple but well-defined structure that can be fully exploited by the powerful decoupling trick. Therefore, based on the well structured IVA framework we propose to develop a powerful framework where identification independent subspaces can be achieved jointly.

**Estimation of Sparsity Levels**  As we presented in Chapter 5, SparseICA-EBM is an attractive ICA algorithm for applications where prior information about the sparsity of the sources is available. However, there are many applications where not all sources possess the same amount of sparsity. This motivates the development of techniques to adaptively estimate optimal values of the parameters $\lambda_n$ and $\epsilon_n$, which is greatly facilitated through the use of the decoupling trick. This would significantly increase the separation performance and improve the quality of the final extracted sources, especially when sources have different levels of sparsity.

### 6.2.2 Applications

Our aim in developing new algorithms is not only to demonstrate the versatility of the ICA/IVA framework but also to consider applications for which these algorithms are ideal. Thus, we propose to use our algorithms for the detection of an abandoned object in a given video and also use them in new applications such as the analysis of genetic data.

**Video surveillance**   Automated detection of abandoned object is an important application in video surveillance for security purposes. Since they minimize the assumptions placed on the data, approaches using BSS techniques such as ICA, have shown to be an attractive way of detecting abandoned objects in a diverse set of scenarios requiring no prior information about the objects. However, the underlying PDFs of the sources can be quite complicated, thus BSS algorithms with more flexible models for the PDFs should provide more accurate detection results. As we discussed in Chapter 3, ICA-EMK has shown superior performance when the PDFs of the underlying sources are complicated, thus motivating the use of ICA-EMK for abandoned object detection.

Although, ICA can be an ideal framework for video surveillance, it can model only one dataset at a time. This may limit its usage to gray-scale frames and ignore the information across the red, green, and blue channels. For this reason, IVA with adaptive MGGD, can be an ideal algorithm for this application.

**Application to genetic data** Association studies based on high-throughput single nucleotide polymorphism (SNP) data have become a popular way to detect genomic regions associated with complex human diseases. SNPs data describes, on a per subject basis, the difference between a nucleotide at a specific location on the genome and a pre-defined genetic template. Since, in general, the deviations from the genetic template are rare, on the order of tens or hundreds, when compared to the gene locations, on the order of hundreds of thousands or millions, the factors underlying SNPs data tend to be sparse. This explains the popularity of sparse PCA methods [75, 77] for the analysis of such data. However, PCA methods are based on the assumption of correlation among the components. This motivates the use of SparseICA-EBM for the analysis of genetic data, since, like sparse PCA methods, SparseICA-EBM enables dynamic control over the degree of sparsity with the natural advantages of ICA over PCA [43].

# Appendix A

# CONVERGENCE OF RA-FP

This section provides the proof of convergence of the RA-FP algorithm, which was presented in Section 4.1.2.

The proof essentially relies on the Riemannian geometry of the space $\mathcal{S}_+^p$, the space of symmetric positive definite, $p \times p$ real matrices [89], [14]. The main geometric property to be used is the *strong convexity of Riemannian distance* [88], which is now explained.

To begin, the length of a differentiable curve $c : [0, 1] \to \mathcal{S}_+^p$ is defined as [89]

$$L(c) = \int_0^1 \|c^{-1}(t)\dot{c}(t)\|_F \times dt, \tag{A.1}$$

where $\|\cdot\|_F$ denotes the Frobenius norm. Let $\mathbf{P}$ and $\mathbf{Q}$ be two points in $\mathcal{S}_+^p$. A curve $c$ is said to connect $\mathbf{P}$ and $\mathbf{Q}$ if $c(0) = \mathbf{P}$ and $c(1) = \mathbf{Q}$. Among all curves connecting $\mathbf{P}$ and $\mathbf{Q}$, there exists a unique curve $\gamma$, whose length is minimum, (recall length is defined by (A.1)). This curve $\gamma$ is called the *geodesic* connecting $\mathbf{P}$ and $\mathbf{Q}$. Its equation, in the notation of (4.11), is [14],[86]

$$\gamma(t) = \mathbf{P}\#_t\mathbf{Q}. \tag{A.2}$$

In particular, this exhibits the geometric meaning of the Riemannian average of $\mathbf{P}$ and $\mathbf{Q}$, defined in Section 4.1.2. The Riemannian average with ratio $t$ of $\mathbf{P}$ and $\mathbf{Q}$ is the point $\gamma(t)$ lying on the geodesic $\gamma$ connecting $\mathbf{P}$ and $\mathbf{Q}$.

Riemannian distance between $\mathbf{P}$ and $\mathbf{Q}$, denoted $d(\mathbf{P}, \mathbf{Q})$ is the length of the geodesic curve $\gamma$, defined by (A.2). Using (A.1), it can be found analytically [89],

$$d(\mathbf{P}, \mathbf{Q}) = \|\log(\mathbf{P}^{-1/2}\mathbf{Q}\mathbf{P}^{-1/2})\|_F, \tag{A.3}$$

The main property of Riemannian distance, used in the proof of convergence of the RA-FP algorithm is its strong convexity [88]. This is defined as follows. Let $\mathbf{R}, \mathbf{P}, \mathbf{Q} \in \mathcal{S}_+^p$ and $\gamma : [0, 1] \to \mathcal{S}_+^p$ the geodesic connecting $\mathbf{P}$ and $\mathbf{Q}$, given by (A.2). Then,

$$d^2(\mathbf{R}, \gamma(t)) \le t\, d^2(\mathbf{R}, \mathbf{Q}) + (1 - t)\, d^2(\mathbf{R}, \mathbf{P})$$

$$-t(1 - t)d^2(\mathbf{P}, \mathbf{Q}). \tag{A.4}$$

This inequality simply means the function $t \mapsto d^2(\mathbf{R}, \gamma(t))$, which is a real-valued function of the real variable $t$, is a strongly convex function.

Consider now, once more, the fixed point equation (4.9). The FP algorithm (4.10), produces iterates $\mathbf{\Sigma}_k$ which converge to the unique fixed point $\hat{\mathbf{\Sigma}}$ of the function $f$, whenever $f$ is contractive. That is, whenever [84]

$$d(f(\mathbf{P}), f(\mathbf{Q})) \le \lambda \times d(\mathbf{P}, \mathbf{Q}) \qquad \lambda < 1 \tag{A.5}$$

for all $\mathbf{P}, \mathbf{Q} \in \mathcal{S}_+^p$. On the other hand, the FP algorithm (4.10) has no guarantee of convergence when $\lambda = 1$, in which case $f$ is said to be non-expansive. Precisely, in this case [84],

$$d(f(\mathbf{P}), f(\mathbf{Q})) \le d(\mathbf{P}, \mathbf{Q}), \tag{A.6}$$

for all $\mathbf{P}, \mathbf{Q} \in \mathcal{S}_+^p$. For function $f$ as defined in (4.6), numerical experiments have shown that, in a neighborhood of the true value $\mathbf{\Sigma}$, this function is contractive when $\beta < 2$, but only non expansive, when $\beta \ge 2$. We numerically support our assumption regarding the non-expanding behavior of the map $f$ by the following experiment.

We generate data with $N = 1000$, $p = 3$, and $\sigma = 0.5$. Then, we constructe

AR(1) covariance matrices $\mathbf{M}_1$ and $\mathbf{M}_2$, with correlation parameters $\sigma_1$ and $\sigma_2$ which range over the interval $(0, 1)$ and then we compute the Riemannian distances $d(\mathbf{M}_1, \mathbf{M}_2)$ and $d(f(\mathbf{M}_1), f(\mathbf{M}_2))$. Fig. A.1 and Fig. A.2 show the difference $d(f(\mathbf{M}_1), f(\mathbf{M}_2)) - d(\mathbf{M}_1, \mathbf{M}_2)$ as a function of $\sigma_1$ and $\sigma_2$. If this difference is non-positive in some region, then $f$ is non-expansive over that region. Note that the results achieved for this set of parameters, are similar to the results achieved when $\sigma = 0.2$ and $\sigma = 0.8$, as well as for different values of the dimension $p$. For $\beta = 4$ and $\beta = 8$, the difference is mostly non-positive, and it is non-positive in the neighborhood of the true value of $\sigma$.

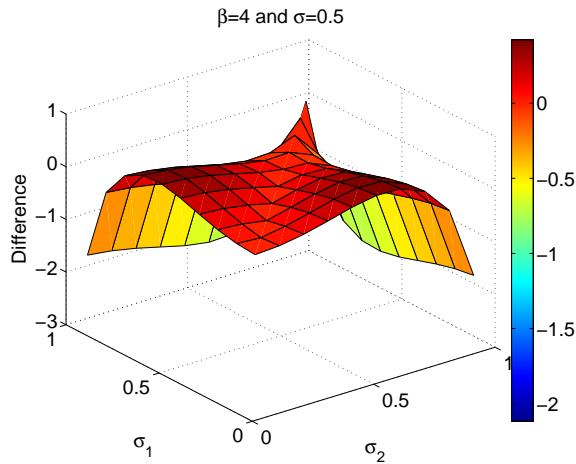

Fig. A.1. Difference between $d(f(\mathbf{M}_1), f(\mathbf{M}_2))$ and $d(\mathbf{M}_1, \mathbf{M}_2)$ as a function of $\sigma_1$ and $\sigma_2$ for $\beta = 4$.

The mathematical explanation of the RA-FP convergence is given in the following proposition.

**Proposition 1.** *Let $f : \mathcal{S}_+^p \to \mathcal{S}_+^p$ be a function, which has a fixed point $\hat{\mathbf{\Sigma}}$. Assume there exists a neighborhood $U$ of $\hat{\mathbf{\Sigma}}$, such that $\hat{\mathbf{\Sigma}}$ is the unique fixed point of $f$ in $U$. Assume also $f$ is non-expansive in $U$. That is, for $\mathbf{P}, \mathbf{Q} \in U$, inequality (A.6) holds. If $\mathbf{\Sigma}_0 \in U$ and, $\mathbf{\Sigma}_{k+1}$ is defined by the RA-FP algorithm (4.13), for $k = 0, 1, 2, \ldots$, then the sequence $\{\mathbf{\Sigma}_k\}$ remains in $U$ and converges to $\hat{\mathbf{\Sigma}}$, as $k \to \infty$.*

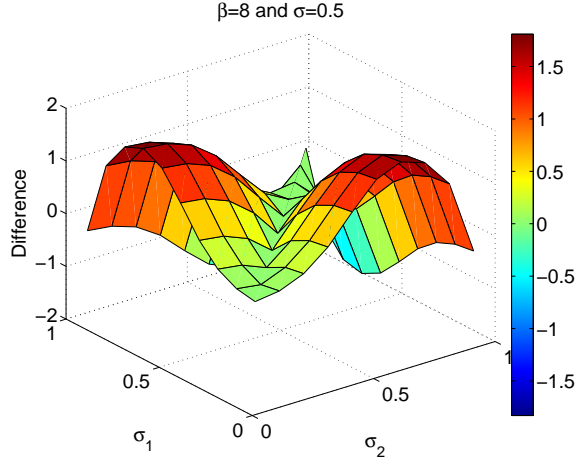

FIG. A.2. Difference between $d(f(\mathbf{M}_1), f(\mathbf{M}_2))$ and $d(\mathbf{M}_1, \mathbf{M}_2)$ as a function of $\sigma_1$ and $\sigma_2$ for $\beta = 8$.

**Proof :** Assume $\mathbf{\Sigma}_k \in U$. Since $\hat{\mathbf{\Sigma}} \in U$, it follows from (A.6),

$$d(f(\hat{\mathbf{\Sigma}}), f(\mathbf{\Sigma}_k)) \leq d(\hat{\mathbf{\Sigma}}, \mathbf{\Sigma}_k).$$

But $\hat{\mathbf{\Sigma}}$ is a fixed point of $f$, so $f(\hat{\mathbf{\Sigma}}) = \hat{\mathbf{\Sigma}}$. Replacing (A.7) in the above inequality, it follows that

$$d(\hat{\mathbf{\Sigma}}, f(\mathbf{\Sigma}_k)) \leq d(\hat{\mathbf{\Sigma}}, \mathbf{\Sigma}_k). \tag{A.7}$$

Now, apply the strong convexity property (A.4), with $\mathbf{R} = \hat{\mathbf{\Sigma}}$, $\mathbf{P} = \mathbf{\Sigma}_k$, $\mathbf{Q} = f(\mathbf{\Sigma}_k)$, and $t = t_k$. Using (4.13) and (A.2), this gives

$$d^2(\hat{\mathbf{\Sigma}}, \mathbf{\Sigma}_{k+1}) \leq t_k\, d^2(\hat{\mathbf{\Sigma}}, \mathbf{\Sigma}_k) + (1 - t_k)\, d^2(\hat{\mathbf{\Sigma}}, f(\mathbf{\Sigma}_k))$$

$$- t_k(1 - t_k) d^2(\mathbf{\Sigma}_k, f(\mathbf{\Sigma}_k)).$$

Replacing (A.7) in this last inequality, it follows after a short calculation

$$d^2(\hat{\mathbf{\Sigma}}, \mathbf{\Sigma}_k) - d^2(\hat{\mathbf{\Sigma}}, \mathbf{\Sigma}_{k+1}) \geq t_k(1 - t_k) d^2(\mathbf{\Sigma}_k, f(\mathbf{\Sigma}_k)). \tag{A.8}$$

This shows that $d(\hat{\mathbf{\Sigma}}, \mathbf{\Sigma}_{k+1}) \leq d(\hat{\mathbf{\Sigma}}, \mathbf{\Sigma}_k)$. So if $\mathbf{\Sigma}_k$ belongs to $U$, so does $\mathbf{\Sigma}_{k+1}$. Thus, if $\mathbf{\Sigma}_0 \in U$, then the sequence $\{\Sigma_k\}$ remains in $U$. To prove this sequence converges to $\hat{\Sigma}$, sum (A.8) over $k = 0, \ldots, n-1$. This gives,

$$d^2(\hat{\Sigma}, \Sigma_0) - d^2(\hat{\Sigma}, \Sigma_n) \geq \sum_{k=0}^{n-1} t_k(1 - t_k)d^2(\Sigma_k, f(\Sigma_k)) \tag{A.9}$$

The right hand side of this inequality is bounded above by $d^2(\hat{\Sigma}, \Sigma_0)$, which does not depend on $n$. Therefore,

$$\sum_{k=0}^{\infty} t_k(1 - t_k)d^2(\Sigma_k, f(\Sigma_k)) < +\infty. \tag{A.10}$$

To complete the proof, take the neighborhood $U$ of $\hat{\Sigma}$ to be compact. This can be done without any loss of generality.

The sequence $\Sigma_k$ converges to $\hat{\Sigma}$ if and only if $d(\hat{\Sigma}, \Sigma_k) \to 0$. It is now shown that assuming this is not true would lead to a contradiction.

By (A.8), the sequence of distances $d(\hat{\Sigma}, \Sigma_k)$ is decreasing. Therefore, if it does not converge to 0, there exists a positive number $\delta$ such that $d(\hat{\Sigma}, \Sigma_k) \geq \delta$ for all $k$.

Let $C$ be the set of matrices $\Sigma$ such that $d(\hat{\Sigma}, \Sigma) \geq \delta$. This is a closed set. Therefore, the set $U \cap C$ is compact. Note the function $\Sigma \mapsto d(\Sigma, f(\Sigma))$ is continuous. Therefore, this function reaches its minimum, say $c$, over $U \cap C$. Since $U \cap C$ does not contain any fixed points of $f$, it follows that $c > 0$.

It has been proved that $\Sigma_k \in U$ for all $k$, and that, assuming $\Sigma_k$ does not converge to $\hat{\Sigma}$, $\Sigma_k \in C$ for all $k$. In this case, $\Sigma_k \in U \cap C$ for all $k$. This implies $d(\Sigma_k, f(\Sigma_k)) \geq c$ for all $k$. Replacing in the right hand side of (A.10),

$$\sum_{k=0}^{\infty} t_k(1 - t_k)d^2(\Sigma_k, f(\Sigma_k)) \geq c^2 \sum_{k=0}^{\infty} t_k(1 - t_k).$$

Since $t_k = \frac{1}{1+k}$, this sum is infinite, which contradicts (A.10).

Since the assumption that $d(\hat{\Sigma}, \Sigma_k)$ does not converge to zero has lead to a contradiction, it follows that $d(\hat{\Sigma}, \Sigma_k) \to 0$, which means that $\Sigma_k$ converges to $\hat{\Sigma}$. ∎

Recall that function $f$ is defined by (4.6) within the ML framework for the estimation of MGGD parameters. As discussed right after (4.9), the maximum likelihood estimate $\hat{\Sigma}$ of the scatter matrix $\Sigma$ is a fixed point of this function. Moreover, as discussed after (A.6), numerical experiments have shown that this function verifies the assumption of non-expansivity in a small neighborhood of the true value, and since for sufficiently large sample size the maximum likelihood estimate $\hat{\Sigma}$ is expected to be close to the true value, Proposition 1 asserts that the RA-FP algorithm (4.13) applied to function $f$ converges to $\hat{\Sigma}$, if it is initialized in a small neighborhood of the true value. This is in full agreement with the numerical results of Section 4.1.3.